\pdfoutput=1
\documentclass[10pt]{article}
\PassOptionsToPackage{dvipsnames}{xcolor}
\usepackage[preprint]{tmlr}

\usepackage[utf8]{inputenc}
\usepackage{microtype}
\usepackage{amsmath, amssymb, amsthm}
\usepackage{amsfonts}
\usepackage{nicefrac}
\usepackage{graphicx}
\usepackage{float}
\newfloat{algorithm}{tbp}{loa}
\floatname{algorithm}{Algorithm}
\graphicspath{{figures/}}
\usepackage{booktabs}
\usepackage{tabularx}
\usepackage{longtable}
\usepackage[dvipsnames]{xcolor}
\providecolor{NavyBlue} {cmyk}{0.94,0.54,0,0}
\providecolor{TealBlue} {cmyk}{0.86,0,0.34,0.02}
\providecolor{RedOrange}{cmyk}{0,0.77,0.87,0}
\providecolor{Magenta}  {cmyk}{0,1,0,0}
\usepackage{caption}
\usepackage{subcaption}
\usepackage[colorlinks=true, linkcolor=NavyBlue, citecolor=NavyBlue, urlcolor=NavyBlue]{hyperref}
\usepackage{url}
\hypersetup{pdftitle={Audited Gradients, Anchored Policies: Model-Based Reinforcement Learning for Efficient Flow Control}, pdfauthor={Ruige Kong, Stefania Fresca}}
\usepackage{multicol}


\newcommand{\loss}{\mathcal{L}}

\newcommand{\E}{\mathbb{E}}
\newtheorem{proposition}{Proposition}

\newif\ifreviewmode
\reviewmodefalse
\ifreviewmode
  
  \newcommand{\reply}[1]{{\color{Magenta}\small\textsf{[reply: #1]}}}
  \newcommand{\SF}[1]{{\color{RedOrange}\small\textsf{[SF: #1]}}}
\else
  
  \newcommand{\reply}[1]{}
  \newcommand{\SF}[1]{}
\fi

\newif\ifsixmode
\sixmodefalse
\ifsixmode
  
\else
  
\fi

\title{Audited Gradients, Anchored Policies: Model-Based\\ Reinforcement Learning for Efficient Flow Control}

\author{\name Ruige Kong \email rk684@uw.edu \\
      \addr Department of Mechanical Engineering, University of Washington, Seattle, WA 98195, United States \\
      NSF AI Institute in Dynamic Systems, University of Washington, Seattle, WA 98195, United States
      \AND
      \name Stefania Fresca \email sfresca@uw.edu \\
      \addr Department of Mechanical Engineering, University of Washington, Seattle, WA 98195, United States \\
      NSF AI Institute in Dynamic Systems, University of Washington, Seattle, WA 98195, United States}

\renewcommand{\floatpagefraction}{0.80}
\begin{document}
\maketitle

\begin{abstract}
Feedback controllers for flows are expensive to design, since each candidate must be evaluated
by numerically solving the governing equations with a high-fidelity solver. Differentiable surrogates reduce that cost, but they also introduce a failure mode that prediction error cannot detect. Policy optimization uses the derivative of the objective with respect to the action, which we call the action gradient and which can be computed either through the surrogate or through the high-fidelity solver. Optimization through the surrogate uses only its action gradient, and a surrogate can follow the flow trajectory closely while misestimating this derivative. We address this failure mode by building controllers from action gradients that have been measured or audited. The surrogate action gradient is compared against a central finite-difference estimate from the high-fidelity solver on held-out states, under acceptance criteria fixed before the comparison, and we call that comparison an audit. A low-dimensional feedback law is then fitted to the surrogate action gradient when the surrogate passes the audit, and to the solver estimate when it does not. This law initializes a neural policy, and a quadratic penalty holds the policy near it during optimization. The proposed approach is validated on four test cases. The resulting controllers outperform or match a model-free baseline trained under the same protocol, while using 3 to 156 times fewer high-fidelity solver interactions across the four test cases. Three findings explain why each part of the procedure is needed. One surrogate predicts the objective at correlation 0.9996 but its action gradient at only 0.459, so prediction accuracy does not identify a usable surrogate. Another surrogate, whose action gradient agrees with the solver estimate at correlation 0.996, makes the objective $0.52\%$ worse when the policy is optimized through it without the anchor penalty and from an initialization not derived from the law, so an accurate gradient alone does not provide a working controller. Finally, a surrogate that passes the audit at one fixed action fails it around the actions the deployed controller applies, so a pass is local and the gradient must be measured where the controller operates.
\end{abstract}

\section{Introduction}

Feedback control selects the action using measurements of the evolving flow, and it can
decrease drag, noise, and mixing without maintaining a strong input at all times
\citep{brunton2015closed,vinuesa2022deep}. Designing such a controller is expensive because every
candidate must be evaluated by a high-fidelity solver that resolves the governing equations.

Control theory and reinforcement learning use different evidence when designing controllers for a
dynamical system such as a fluid flow \citep{vernade2026foundations}.
Control theory seeks stability or safety evidence before deployment, while reinforcement learning
normally reports performance after training. This difference matters when the controller is
trained by differentiating through a learned surrogate, because the update depends on the
derivative of the predicted objective with respect to the action, a quantity that post-training performance never examines.

Deep reinforcement learning has established that neural policies can control fluid flows from
sensor measurements, ranging from laminar cylinder wakes \citep{rabault2019artificial} to turbulent
channels \citep{guastoni2023deep} and separation bubbles \citep{font2025bubble}. However,
model-free training may require tens of thousands of high-fidelity solver interactions
\citep{rabault2019artificial,paris2021robust}.
Model-based methods reduce this cost by learning a surrogate and optimizing the controller
with predicted trajectories, and in pathwise methods, used in this work, the policy is updated with the derivative of the predicted objective with respect to the action \citep{mora2021pods,liu2024adjoint}.

A controller trained by differentiation employs the derivative of the objective with respect to the action. We call this derivative the \textit{action gradient}, whether it is computed through the surrogate by automatic differentiation of the predicted objective, or through the high-fidelity solver by finite differences. A surrogate can
predict a trajectory accurately and still predict the response to an action perturbation
incorrectly. Therefore, prediction error does not establish action gradient accuracy.
Objective-aware model learning addresses this by reweighting the prediction errors according to their effect on control
\citep{lambert2019objective,voelcker2023vagram}.
Methods for chaotic systems address a different part of the problem, namely how an
unreliable gradient is used during optimization
\citep{metz2021gradients,suh2022differentiable,gao2024adaptive}. The present study adds an
independent measurement of the action gradient from the high-fidelity solver before controller
optimization.

These observations determine the evaluation procedure used in this work. Prediction accuracy and action gradient accuracy are evaluated separately because they measure different properties of the surrogate.
The gradient comparison is local to the sampled states and actions, so it must be repeated around the actions the deployed controller applies. The comparison also requires high-fidelity simulation, so every trajectory used by the comparison is included in the simulation cost.

\subsection{Contributions of this work}

\begin{figure}[!tbp]
\centering
\includegraphics[width=\textwidth]{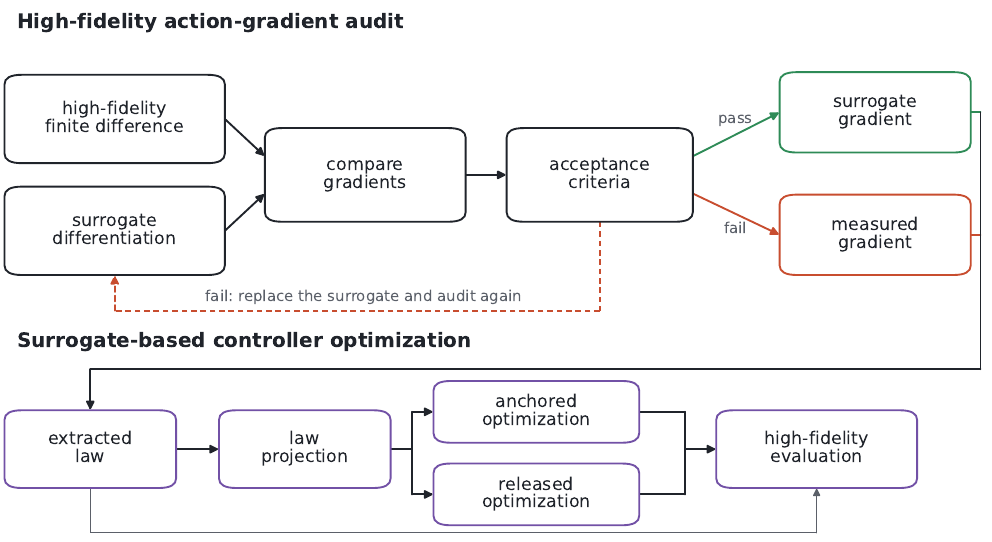}
\caption{Surrogate action gradients are compared with high-fidelity central differences on held-out states under fixed acceptance criteria. After a pass, the extracted law is fitted to the surrogate gradient, which is inexpensive to evaluate. After a failure instead, it is fitted to the measured gradient, unless a replacement surrogate passes a new audit. The extracted law is transferred to the neural policy by
law projection, and used as the reference of the anchor penalty, which a cone around the law replaces in the case of a failure. The law, the anchored controller and the released controller are each evaluated in the solver. Black boxes refer to the audit, green and orange paths to its two outcomes,
and purple boxes to the controller construction.}
\label{fig:method-schematic}
\end{figure}

In this work, we propose an audited anchor-based pathwise controller, that is, a controller trained
by differentiating through a learned surrogate whose action gradient has been measured against the
high-fidelity solver, and held during training near a feedback law extracted from that measurement. Figure~\ref{fig:method-schematic} shows a schematic of the proposed approach.

The first contribution is the \textit{action gradient audit}, defined as a comparison
between surrogate derivatives and central finite differences computed with the high-fidelity
solver on held-out states. Here, held-out states are states excluded from surrogate fitting.
The comparison is decided by acceptance criteria, three statistics measuring the agreement
between the surrogate gradient and the solver estimate, and each is compared with a threshold fixed
before controller optimization. The surrogate passes the audit only when all three statistics meet
their thresholds. Section~\ref{sec:threshold} relates the measured gradient alignment to the
standard descent condition for an approximate gradient.

The second contribution is a controller-construction strategy based on the measured derivative field. Within this strategy, an \textit{extracted law} is a low-dimensional parametric feedback law fitted to audited descent directions, those measured with the solver or those obtained by means of a surrogate that has passed the audit. Consequently, the inexpensive surrogate gradient is used wherever the audit shows it can be trusted, and the measured gradient keeps controller construction possible where it cannot. This law is transferred to a neural policy by matching the
actions the two produce, and we call this step \textit{law projection}. During
pathwise optimization, an \textit{anchor penalty} provides quadratic regularization toward the
extracted law. The weight of the anchor penalty is either held fixed or reduced linearly to zero during training, and the two resulting controllers, called anchored and released, are evaluated separately.

The third contribution consists of an empirical assessment of the action gradient audit and of
computational cost across four flow control test cases. The experiments compare prediction
accuracy with action gradient accuracy. They also measure the extent to which the comparison depends on
the action around which the derivative is taken. The high-fidelity finite-difference computations behind each reported controller are counted in the model-based simulation cost.

\subsection{Related works}

The prior work relevant to this study falls into four areas: reinforcement learning for flow control, objective-aware model learning, pathwise policy optimization, and the construction of feedback laws from measured gradients.

Reinforcement learning for flow control provides the general context. Neural policies have been demonstrated for laminar cylinder wakes, confined bluff body wakes, turbulent channels, and turbulent separation bubbles \citep{rabault2019artificial,zhang_2024_surrogate_drl_square_cylinder,guastoni2023deep,zhou2025channel,font2025bubble}. A broad survey is given by \citet{vinuesa2022deep}. Model-based reinforcement learning reduces the high-fidelity simulation cost by learning a dynamics surrogate and optimizing the policy with predicted trajectories \citep{chua2018deep,janner2019when,hafner2020dream,langlois2019benchmarking,weiner_2024_model_based_drl}. These methods assess the learned model through state prediction error or objective prediction error.

Objective-aware model learning shows why prediction error is not a sufficient criterion for accepting a surrogate. A prediction loss can emphasize errors that have little effect on the objective and underweight errors that strongly affect the policy update \citep{lambert2019objective}. VaGraM addresses this mismatch by weighting state prediction errors with the gradient of a learned value function \citep{voelcker2023vagram}, and studies of machine learned molecular force fields report a related result, in which low prediction error still produces unstable or unphysical trajectories \citep{fu2023forces}. These studies show that the quantity used downstream should be evaluated directly. In the present work, that quantity is the action gradient, and it is evaluated directly against the high-fidelity solver instead of through a reweighted training loss. The two approaches are complementary because a surrogate trained with an objective-aware loss can itself be audited before it is used for control.

Pathwise policy optimization, the third area, updates the policy by differentiating the predicted objective through the surrogate, and each update depends on the action gradient. A differentiable simulator can supply the action derivative directly \citep{mora2021pods}, but long derivative chains have high variance in chaotic systems and can produce directions that fail to decrease the physical objective \citep{parmas2018pipps,suh2022differentiable,metz2021gradients}. Existing methods address these difficulties by differentiating governing equations with learned terms \citep{liu2024adjoint}, adding terminal value estimates to short rollouts \citep{amos2021modelbasedsvg}, or switching between simulator and critic gradients when the analytic gradient degrades \citep{gao2024adaptive}. The present method instead verifies the gradient before optimization begins, using short pathwise rollouts, and builds the controller from the surrogate derivative, if it passes an independent high-fidelity comparison, and from the solver estimate otherwise.

A verified derivative field must then be converted into a feedback controller. Low-dimensional feedback laws have a long history in fluid mechanics, including reduced-order and sensitivity-based rotary control of cylinder wakes \citep{bergmann_2005_pod_rotary,sato_2025_parametric_pod}. SINDy-RL distills a trained neural policy into an interpretable symbolic policy \citep{zolman_2024_sindy_rl}. The extracted law used in this work is instead fitted before any policy training, directly to descent directions measured with high-fidelity finite differences. Gradient-based policy supervision also appears in diffusion policies and Q-score matching \citep{yang2023dipo,psenka2024qsm}, where the supervising gradient comes from a learned critic, and policy anchoring appears in behavior-regularized offline reinforcement learning \citep{fujimoto2021minimalist}, where the reference is the behavior data. %
In the present approach, the anchor is derived from the measured derivative field.

Relative to these four areas, and to the best of our knowledge, the present work is the first active flow control study that compares the surrogate action gradient with a high-fidelity finite-difference estimate before controller construction, measures the extent to which that comparison depends on the sampled states and actions, and includes the cost of the comparison in the simulation cost of the model-based approach. It is also the first, to our knowledge, in which the feedback law that initializes and regularizes the policy is fitted only to audited gradients, instead of to an unverified learned quantity.

\section{Problem formulation}
\label{sec:background}

Two systems of partial differential equations define the dynamics of the four test cases analyzed in this work. The equations are stated first, followed by the common control notation and the evaluation quantities.

\subsection{Governing equations}
\label{sec:governing}

Every controller is evaluated by direct numerical simulation, so the governing equations are resolved without a turbulence model. Consequently, each controller score requires a complete high-fidelity rollout, and each central finite-difference estimate requires two rollouts over the audit horizon $H$, defined in Section~\ref{sec:audit}. Section~\ref{sec:cost} {includes both} in the simulation cost.

\paragraph{Incompressible Navier Stokes equations.} Three test cases use the incompressible
Navier Stokes equations. These are the rotating cylinder wake in Section~\ref{sec:latentfail}, the
three cylinder cluster in Section~\ref{sec:repair}, and forced two-dimensional turbulence in
Section~\ref{sec:kolmogorov}. They are written for a velocity field
$\mathbf{u}$ and a pressure field $p$, both functions of position and time,
\begin{equation}
    \frac{\partial \mathbf{u}}{\partial t}
    +(\mathbf{u}\cdot\nabla)\mathbf{u}
    =-\nabla p+\frac{1}{Re}\nabla^{2}\mathbf{u}
    {}+\mathbf{f},
    \qquad
    \nabla\cdot\mathbf{u}=0.
    \label{eq:navier-stokes}
\end{equation}
The equations are written in dimensionless form, so the Reynolds number $Re$ measures the ratio of
inertial to viscous effects. In the two cylinder cases the reference scales are the cylinder
diameter and the inflow velocity, and in the turbulence case the box has side $2\pi$ and the fixed
forcing has unit amplitude in these units. For the rotating cylinder and the three cylinder
cluster, $\mathbf f=\mathbf 0$ and the actions set the tangential velocities imposed at the cylinder
walls. For forced two-dimensional turbulence, the domain is periodic and the action sets four
amplitudes of a volume force. In the velocity formulation this force takes the form
\begin{equation}
    \mathbf f(y,t)=\left[\sin(k_fy)+\sum_{k=4}^{7}a_k(t)\sin(ky)\right]\mathbf e_x,
    \qquad k_f=4,
    \label{eq:kolmogorov-force}
\end{equation}
where the first term is the fixed forcing that sustains the turbulence and the four actions set
the amplitudes of the control modes, the first of which shares the wavenumber $k_f$ of the fixed
forcing, so that $a_4$ rescales the forcing itself. Numerically, the solver advances the equivalent
vorticity equation on a doubly periodic square of side $2\pi$. This equation is obtained by taking
the curl of the momentum equation in Equation~\eqref{eq:navier-stokes} with the forcing of
Equation~\eqref{eq:kolmogorov-force}, and in two dimensions the curl has the single component
$\omega=(\nabla\times\mathbf u)\cdot\mathbf e_z$, the vorticity. The divergence-free velocity is
recovered from the vorticity through a stream function whose spatial mean is set to zero, so the
flow has no mean velocity.

\paragraph{Kuramoto--Sivashinsky equation.} The parametric chaotic Kuramoto--Sivashinsky equation for a
field $u(x,t)$ on a periodic interval of length $L$, with $L=22$, takes the form
\begin{equation}
    \frac{\partial u}{\partial t}
    +u\frac{\partial u}{\partial x}
    +\frac{\partial^{2} u}{\partial x^{2}}
    +\frac{\partial^{4} u}{\partial x^{4}}
    =\sum_{j=1}^{8} a_{j}(t)\,\chi_{j}(x)
    +\mu\cos\!\left(\frac{4\pi x}{L}\right).
    \label{eq:ks-pde}
\end{equation}
The eight actions $a_{j}$ act through fixed Gaussian
profiles $\chi_{j}$ centered at equally spaced positions, so the control acts at eight fixed
locations. The parameter $\mu$ sets the amplitude of a stationary forcing that breaks the
translational symmetry of the equation, and changing $\mu$ changes the dynamical regime in which
the controller operates. The evaluated forcing amplitudes are $\mu=0.225$ and $\mu=0.121$,
following the benchmark of \citet{botteghi2024parametric}. Equation~\eqref{eq:ks-pde} reproduces the implementation of that benchmark, including the sign of the forcing term.

Table~\ref{tab:flow-overview} and Appendix~\ref{app:surrogate} report the Reynolds number of each Navier Stokes case and the resolution used in each high-fidelity simulation. The four cases
span boundary and volume actuation, wall-bounded and periodic domains, and one- to
eight-dimensional action spaces.

\subsection{Control problem and evaluation quantities}

The four test cases use the same discrete-time notation, but each one has its own objective, and the objectives are on different scales. Let $\mathbf x_t$ denote the high-fidelity state and $\mathbf a_t$ the action applied at step $t$, %
\begin{equation*}
    \mathbf x_{t+1}=F(\mathbf x_t,\mathbf a_t),
    \qquad
    \mathbf o_t=h(\mathbf x_t),
    \qquad
    \mathbf a_t=\pi(\mathbf o_t).
\end{equation*}
The map $F$ applies the action $\mathbf{a}_t$ and integrates the governing equations over one control step, which spans several time
steps of the high-fidelity solver. The index $t$ counts control steps, and one control step is one interaction with the high-fidelity solver, the unit employed in Section~\ref{sec:cost} to compute the simulation cost. The map $h$ returns the observation $\mathbf o_t$ available to the controller.
In the parametric chaotic equation the observation also contains the forcing amplitude $\mu$, and
Gaussian noise is added to the sensor values that form the controller input, as in the original
benchmark. The policy $\pi$ maps the observation to the action. The surrogate-based controllers rely on a learned model in one of two
forms, as detailed in Section~\ref{sec:training}. The first is a transition model $\widehat F \approx F$ that advances either the state or a latent representation of it. The second
predicts the action gradient directly, without a transition model. In the following, we write $\widehat F$ as $\widehat F_{\boldsymbol\psi}$ when we want to highlight its dependence on the trainable parameters $\boldsymbol\psi$.

In the Navier Stokes test cases, each controlled evaluation is paired with an uncontrolled rollout from the same initial state, so the difference reflects the controller alone and not the choice of initial condition. For objectives reported as a percentage reduction,
let $J_i(\pi)$ denote the value of the objective over the evaluation window for policy $\pi$ and
initial condition $i$, and let $J_i(0)$ denote the same quantity for the uncontrolled rollout, in
which the action is zero. The paired gain is
\begin{equation}
    G_i(\pi)=100\frac{J_i(0)-J_i(\pi)}{J_i(0)}.
    \label{eq:paired-gain}
\end{equation}
A positive value of $G_i$ indicates that the controller reduces the objective. For each result, we state
whether the reported uncertainty is computed across initial conditions, training seeds, or both.
The parametric chaotic equation is the exception to the previous percentage convention. As in the published benchmark, its results are reported as the state cost $c_1$, defined in Section~\ref{sec:chaotic} as the mean square of the field, accumulated over the episode, for which
lower values are better.

\section{Proposed method}
\label{sec:method}

Controller construction begins with two estimates of the same action gradient, one obtained from
the surrogate by automatic differentiation and one obtained independently from the high-fidelity
solver by finite differences. The audit, a comparison of the two estimates on held-out states under fixed acceptance criteria, decides which of them may be used to construct the controller. When the audit passes, the surrogate gradient, which is inexpensive to evaluate at any state, is used. When it fails, either a replacement surrogate must pass a new audit or the solver estimate alone defines the controller, and the surrogate may then adjust the action only within a narrow constraint around the measured descent direction. In both cases {the audited descent directions} are used to construct a low-dimensional feedback law. This
law initializes a neural policy and serves as the reference of the anchor penalty during pathwise
optimization. The performance of the final controller is evaluated with the high-fidelity solver.

\subsection{Action gradient audit}
\label{sec:audit}

The action gradient of a learned surrogate is compared with an estimate computed by the
high-fidelity solver on the same states, around the same action, and over the same horizon. Let
$J_H(\mathbf x,\mathbf a)$ denote the control objective accumulated over a horizon $H$ from state
$\mathbf x$ under the \textit{reference action} $\mathbf a$, around which the derivative is taken.
On the rotating cylinder and in forced turbulence, $\mathbf a$ is applied unchanged at every step, so the derivative is taken with respect to a constant action. In the parametric chaotic equation $\mathbf a$ is the action of a fixed base controller, introduced in Section~\ref{sec:chaotic}, which we call the fixed reference action, and it is applied at the first step only, and the remaining $H-1$ actions are produced by the same controller, because the derivative of a 50-step objective with respect to the first action is defined only along the closed-loop trajectory that follows it. Section~\ref{sec:chaotic} describes such a controller and the correction built on the audited gradient. In the three cylinder cluster, $H=1$. For action coordinate $i$,
the corresponding derivative is estimated by two high-fidelity rollouts from the same state and over the same horizon
\begin{equation}
 g_i^{\rm HF}(\mathbf x,\mathbf a,H)=
 \frac{J_H(\mathbf x,\mathbf a+\epsilon \mathbf e_i)-J_H(\mathbf x,\mathbf a-\epsilon \mathbf e_i)}{2\epsilon},
 \label{eq:paired-gradient}
\end{equation}
where $\mathbf e_i$ is the unit vector of coordinate $i$ and the two rollouts differ only in the sign of the perturbation $\epsilon \mathbf e_i$. %

The surrogate predicts the accumulated objective $\widehat J_H$ as a differentiable function of
$\mathbf a$ under the same convention as the solver, and automatic differentiation gives
\begin{equation}
 \widehat g_i(\mathbf x,\mathbf a,H)=\frac{\partial \widehat J_H(\mathbf x,\mathbf a)}{\partial a_i}.
 \label{eq:surrogate-gradient}
\end{equation}
The audit compares $\mathbf g^{\mathrm{HF}}$ and $\widehat{\mathbf g}$ on the held-out states, using three statistics. The first is the pooled correlation, the Pearson correlation between the two gradients, computed over all held-out states and all action components together. The second is the sign agreement, the fraction of the entries at which the two gradients have the same sign. The third is the median, over actuators, of the correlation between $g_i^{\rm HF}$ and $\widehat{g}_i$ across the held-out states. Here, each component $a_i$ of the action commands one actuator, one physical control device. For example, in the three cylinder cluster, each cylinder rotates independently and the action has three components.

We write $\tau_1$, $\tau_2$, and $\tau_3$ for the thresholds applied to the pooled correlation, the
sign agreement, and the median actuator correlation, and the surrogate gradient is used to construct the controller only when each statistic meets its threshold. All test cases use $\tau_1=0.70$,
$\tau_2=0.70$, and $\tau_3=0.50$, where the third value is lower because each per-actuator
correlation is computed from fewer samples. The sampled states, reference actions, rollout horizons,
perturbation sizes, and thresholds are all fixed before each audit. For the three cylinder and chaotic equation audits, no decision changes when $\tau_1$ is varied from $0.50$ to $0.90$, $\tau_2$ from $0.50$ to $0.89$, or $\tau_3$ from $0.30$ to $0.63$, with the other two held fixed. These thresholds are empirical, and no theoretical result links them to the performance of a trained controller. Appendix~\ref{app:thresholds} compares the median actuator correlation with the minimum, and the pooled correlation with the per-state cosine of Section~\ref{sec:threshold}. It also shows that no audit decision depends on the finite-difference step $\epsilon$. The rotating cylinder has one actuator, and its audit uses only the first two statistics. The forced turbulence surrogate fails all three criteria, with pooled correlation $0.53$, sign agreement $0.55$, and median actuator correlation $-0.34$, and Section~\ref{sec:kolmogorov} reports how its controller was built from the measured gradient field alone and checked in the solver.

\paragraph{State dependence of the measured action gradient.}

The finite-difference evaluations also quantify how the measured gradient varies across states. Let
$\mathbf g_s = \mathbf g^{\mathrm{HF}}(\mathbf x_s,\mathbf a,H)$ be the high-fidelity gradient at held-out state $\mathbf x_s$, for $s=1,\ldots,S$, and let $\bar{\mathbf g}$ be their mean. The state-independent fraction
\begin{equation}
    f=\frac{S\,\lVert \bar{\mathbf g}\rVert^{2}}{\sum_{s}\lVert \mathbf g_s\rVert^{2}},
    \qquad
    \bar{\mathbf g}=\frac1S\sum_s \mathbf g_s ,
    \label{eq:feedback-content}
\end{equation}
equals one when the gradient is the same vector at every sampled state, so that the descent direction does not depend on the state, and zero
when its state average vanishes. Values near one mean that a single direction in action space carries most of the energy of the measured
field. Two caveats limit what this statistic can support. First, $f$ summarizes a first-order response over one horizon, while a controller
is evaluated over a complete rollout, so a value near one does not
imply that a fixed action controls the flow effectively over a full rollout. Second, $f$ is
measured only at the sampled states and reference actions, so it says
nothing about the states and actions a trained controller visits later.
Appendix~\ref{app:feedback-content} reports $f$ for each test case, together with the action coordinates and reference actions used.

\subsection{Descent condition for an approximate action gradient}
\label{sec:threshold}

The descent lemma \citep{beck2017first,nesterov2018lectures} provides a local condition under which an update of the action along the surrogate gradient, taken at a fixed state $\mathbf{x}$ with step size $\eta>0$, decreases the high-fidelity objective $J_H$.
Let $\mathbf g=\nabla_{\mathbf a}J_H(\mathbf x,\mathbf a)$ and $\widehat{\mathbf g}=\nabla_{\mathbf a}\widehat J_H(\mathbf x,\mathbf a)$ and assume that %
$\mathbf g(\mathbf{x}, \cdot)$ is
$L$-Lipschitz along the segment from $\mathbf a$ to
$\mathbf a-\eta\widehat{\mathbf g}$, and that this segment lies within the admissible action space. The descent lemma
then gives
\begin{equation}
    J_H(\mathbf x,\mathbf a-\eta\widehat{\mathbf g})-J_H(\mathbf x,\mathbf a)
    \;\le\;
    -\eta\,\mathbf g^\top\widehat{\mathbf g}
    +\frac{L\eta^2}{2}\lVert\widehat{\mathbf g}\rVert^2 .
    \label{eq:local-descent-bound}
\end{equation}
Consequently, the update $\mathbf a^+=\mathbf a-\eta\widehat{\mathbf g}$ decreases the high-fidelity objective when
\begin{equation}
    \mathbf g^\top\widehat{\mathbf g}>0
    \quad\text{and}\quad
    0<\eta<\frac{2\mathbf g^\top\widehat{\mathbf g}}
    {L\lVert\widehat{\mathbf g}\rVert^2}.
    \label{eq:local-descent-condition}
\end{equation}
The first condition in Equation~\eqref{eq:local-descent-condition} means that the angle between the two gradient vectors has to be acute. Moreover, the admissible range of $\eta$ is nonempty whenever the first condition holds, so that a
sufficiently small descent step always exists, and a positive rescaling of $\widehat{\mathbf g}$ only
rescales that range. For this reason, the audit statistics measure directional and sign agreement and disregard the magnitude of the gradient.
They remain empirical acceptance criteria because the smoothness constant $L$ is not estimated.

Let $\theta$ denote the angle between the measured gradient $\mathbf g$ and the surrogate gradient $\widehat{\mathbf g}$, so that $\cos\theta$ equals one when the two point in the same direction and zero when they are orthogonal. The following proposition shows that the bound of Equation~\eqref{eq:local-descent-bound} certifies, for a step along $\widehat{\mathbf g}$, a fraction $\cos^{2}\theta$ of the decrease certified for a step along $\mathbf g$.

\begin{proposition}[Directional efficiency of the quadratic bound]
\label{cor:retained-descent}
Let the assumptions above hold for every step size $\eta>0$ considered, and let
$\mathbf g^\top\widehat{\mathbf g}>0$. For a step $\mathbf a-\eta \mathbf d$ along a direction
$\mathbf d\in\{\widehat{\mathbf g},\,\mathbf g\}$, denote by $\Delta_{\mathbf d}$ the largest decrease of
$J_H$ certified by the bound of Equation~\eqref{eq:local-descent-bound}, that is, the negative of its
right-hand side minimized over $\eta>0$. Then
\begin{equation*}
\Delta_{\widehat{\mathbf g}}=\frac{\lVert \mathbf g\rVert^{2}\cos^{2}\theta}{2L},
\qquad
\Delta_{\mathbf g}=\frac{\lVert \mathbf g\rVert^{2}}{2L},
\qquad
\frac{\Delta_{\widehat{\mathbf g}}}{\Delta_{\mathbf g}}=\cos^{2}\theta,
\end{equation*}
where $\cos\theta =\frac{\mathbf g^\top\widehat{\mathbf g}}{\lVert \mathbf g\rVert\,\lVert\widehat{\mathbf g}\rVert}$. In particular, the ratio is independent of $L$ and of
$\lVert\widehat{\mathbf g}\rVert$, and retaining a fraction
$\rho\in(0,1]$ of the exact-gradient bound requires
$\cos\theta\ge\sqrt{\rho}$.
\end{proposition}

Proposition~\ref{cor:retained-descent} concerns a single action update at a single state, and it does not predict the performance of a trained controller: the Lipschitz constant $L$ is not estimated, and the alignment is known only at the held-out states and reference actions, so it remains unknown along the trajectories the optimizer visits. The proposition only serves as an interpretation of the size of an alignment value. For example, an alignment of $\cos\theta=0.70$ retains $0.70^2\approx 0.49$ of the decrease guaranteed with the exact gradient. %
Appendix~\ref{app:proofs} reports the proof.

\subsection{Surrogate training and controller optimization}
\label{sec:training}

This subsection describes the proposed procedure from surrogate fitting to the evaluation of the trained controller, starting with the data splitting. 

The surrogate is trained on
transition data produced by the high-fidelity solver. The audit is informative about the reliability of the
surrogate only when it is computed on held-out states. Thus, the fitting states and the held-out states
are separated before any training begins. All samples originating from one high-fidelity state remain
on the same side of that division, including the two perturbed responses in
Equation~\eqref{eq:paired-gradient}.%

The surrogate is trained with a loss of two terms. The first is the multi-step prediction loss $\mathcal{L}_{\mathrm{pred}}$, the error of
the surrogate in reproducing the high-fidelity trajectories. Since pathwise optimization also uses the derivative of the predicted
objective with respect to the action, the second term penalizes the
disagreement between that derivative and the gradient measured through
Equation~\eqref{eq:paired-gradient},
\begin{equation}
  \mathcal{L}_{\mathrm{model}}
  = \mathcal{L}_{\mathrm{pred}} + \lambda_g \mathcal{L}_{\mathrm{grad}},
\label{eq:model-loss}
\end{equation}
where $\loss_{\rm grad}$ measures that disagreement at the states where the measured gradient is
available, and $\lambda_g$ weights the second term against the first. Training a network on target values together with their derivatives is known as Sobolev training \citep{czarnecki2017sobolev}. Equation~\eqref{eq:model-loss} applies this idea to the action gradient, and its supervised data come from the solver through Equation~\eqref{eq:paired-gradient}. Because
$\loss_{\rm grad}$ requires supervised data, it is used on the rotating cylinder{,} the three cylinder
cluster and the parametric chaotic equation, whose finite-difference data sets provide labeled data for fitting as well as for the audit. Appendix~\ref{app:surrogate} reports the gradient labels used for fitting.

The gradient term $\loss_{\rm grad}$ does not replace the audit. Both compare the surrogate action gradient with labels of the same kind, but $\loss_{\rm grad}$ is computed on the labeled states used for fitting while the audit is computed on held-out states. Therefore, the audit measures whether the surrogate reproduces a gradient absent from its fitting data. A matched ablation on the rotating cylinder confirms that the distinction matters, %
as Appendix~\ref{app:surrogate} reports. %

\begin{figure}[ht]
\centering
\includegraphics[width=\textwidth]{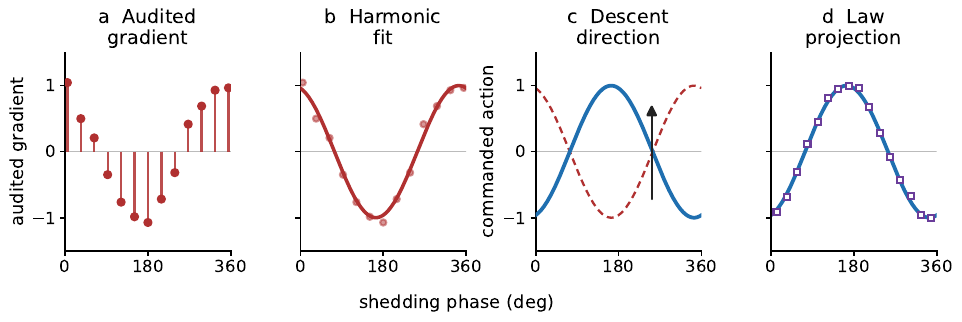}
\caption{The extracted law for the rotating cylinder is a single harmonic of the shedding phase, and the neural policy reproduces it after law projection. Panel a shows the {audited} action gradient at each sampled state, plotted against the shedding phase. %
Panel b shows the single harmonic fitted to those values. Panel c shows the extracted law, where the descent direction equals the fitted harmonic with its sign reversed, and the dashed curve repeats the fitted harmonic for comparison. Panel d overlays the neural policy after law projection (purple squares) on the extracted law (blue).}
\label{fig:law-extraction}
\end{figure}

The gradients measured for the audit serve a second purpose: they can define the extracted law, whether or not the surrogate passes. The
extracted law is a parametric feedback law whose function class is
fixed before controller evaluation. At each held-out state $\mathbf x_s$, $s=1,\ldots,S$, the high-fidelity measurement $\mathbf g_s=\mathbf g^{\mathrm{HF}}(\mathbf x_s,\mathbf a,H)$ introduced with Equation~\eqref{eq:paired-gradient} defines the local descent direction $-\mathbf g_s$, and the parameters defining the direction of the law are chosen by least squares to align its output with these directions across the held-out states. This fit reuses the finite
differences already computed for the audit and requires no additional high-fidelity simulation. The
held-out states are excluded from surrogate fitting but not from the law fit: {a law fitted in this way} is defined from the measured gradient alone, so the surrogate receives no information from the states on which it is evaluated. Moreover, when the surrogate passes the audit, the law fit can be constructed using the action gradient of the surrogate, which can be evaluated at many more states at low computational cost. When the surrogate fails, the law is fitted to the measured gradient alone. We call the gradient used in either way the audited gradient, and Section~\ref{sec:numerical} states which one each test case uses. The amplitude of the law is set separately from this fit, and Appendices~\ref{app:cylinder} and~\ref{app:pinball} give the details. %
For the rotating cylinder, the law class is a single harmonic of the shedding phase, the phase of the vortex shedding cycle defined in Section~\ref{sec:latentfail}, and the fit determines the phase offset of the harmonic, while the amplitude is fixed in advance at $\alpha=0.2$, a bound on the commanded rotation chosen before any high-fidelity evaluation. Figure~\ref{fig:law-extraction} shows a schematic of this construction. The law class is the one choice of the procedure that depends on the flow under investigation. The rotating cylinder uses a physical variable, the shedding phase, while on the three cylinder cluster and the parametric chaotic equation the law is a linear map from the measured forces or sensor values to the actions. On forced turbulence the measured gradient constrains the policy class instead, since the controller admits only actions within a cone around the extracted law set by the measured gradient.

The neural policy is built by performing two steps: law projection and pathwise
optimization. The extracted law has fewer parameters than the neural policy $\pi_{\boldsymbol\phi}$.
Therefore, the law is transferred to the policy by matching the outputs they produce instead of
through their parameters. 
To this end, observations are sampled from
the states used for policy training, the extracted law is evaluated on each of them to produce a target action $\mathbf{a}_{\mathrm{law}}(\mathbf{o})$, and the
policy parameters $\boldsymbol\phi$ are chosen by least squares so that $\pi_{\boldsymbol\phi}(\mathbf{o})$ reproduces these targets. The step requires that every
variable the law depends on be a function of the observation the policy receives, otherwise no choice of $\boldsymbol\phi$ can reproduce the law. For the rotating cylinder, the law depends only on the shedding phase, and the policy observation determines this phase.

Law projection provides the initial policy, and pathwise optimization then improves it by differentiating a finite-horizon objective through the surrogate. The policy loss is%
\begin{equation}
    \loss_\pi(\boldsymbol\phi)
    =
    \E\left[
    \widehat J_H(\pi_{\boldsymbol\phi})
    +\frac{\lambda_a}{H}\sum_{k=0}^{H-1}\lVert\mathbf a_k\rVert^2
    +\frac{\lambda_{\rm anchor}}{H}\sum_{k=0}^{H-1}
    \lVert\pi_{\boldsymbol\phi}(\mathbf o_k)-\mathbf a_{\rm law}(\mathbf o_k)\rVert^2
    \right],
    \label{eq:pathwise-loss}
\end{equation}
where the predicted states follow $\widehat{\mathbf x}_{k+1}=\widehat F_{\boldsymbol\psi}(\widehat{\mathbf x}_k,\mathbf a_k)$, $\mathbf a_k=\pi_{\boldsymbol\phi}(\mathbf o_k)$ is the action selected by the policy at the predicted observation $\mathbf o_k=h(\widehat{\mathbf x}_k)$, and the expectation
is taken over the initial states used for training. No other sources of randomness enter the loss: the surrogate parameters are frozen, the policy is deterministic, and the training seed sets only the random initial weights of $\boldsymbol\phi$, the samples drawn during law projection, and the order in which initial states are drawn. The first term is the objective predicted by the surrogate over the horizon $H$, and the second penalizes the mean squared norm of the action over
the horizon with weight $\lambda_a$.
The third term is the anchor penalty, which penalizes the squared distance between the action
selected by the policy and the action $\mathbf a_{\rm law}$ that the extracted law commands at the same observation. Its weight $\lambda_{\rm anchor}$ sets how closely the policy is held to the law. Offline model-based methods penalize the disagreement among the members of a surrogate ensemble, using it as a proxy for model error
\citep{yu2020mopo}. Equation~\eqref{eq:pathwise-loss} contains no such
term: an inaccurate action gradient can be shared by every member of an ensemble trained on the same data, and a shared error produces no
disagreement for the penalty to detect. The comparison with the high-fidelity solver {can detect} such an error, and for this reason the audit {is used here in place of} the ensemble term.

The weight $\lambda_{\rm anchor}$ is either held at its initial value or reduced linearly to zero over the course of training. The first schedule produces the anchored controller, and the second gives the released controller. The two controllers are built for different reasons, whether holding the policy near the extracted law already yields a good controller and whether the optimizer can be trusted once the anchor is removed. Each is evaluated on held-out initial conditions, distinct from any condition used during training or selection. Algorithm~\ref{alg:procedure} outlines the steps of the proposed approach and marks the points at which the high-fidelity solver is used.
Appendix~\ref{app:surrogate} reports the model and policy architectures, the optimization settings, and the data splits.

\begin{algorithm}[!tbp]
\caption{The audit decides which action gradient constructs the controller: the surrogate gradient after a pass, and the measured gradient after a failure, unless a replacement surrogate passes a new audit on states the failed audit did not use. Besides the fitting transitions of step 1, the high-fidelity solver is used only to measure action gradients in steps 3 and 12, to select the amplitude of the law in step 8 when it is selected in the solver, and to evaluate controllers in steps 11 and 12. Steps 9 and 10 require no solver trajectory.}
\label{alg:procedure}
\small
\begin{tabular}{@{}r@{\hspace{1em}}p{0.93\textwidth}@{}}
\multicolumn{2}{@{}p{\textwidth}@{}}{\textbf{Input:} solver transitions for fitting, states held out from the
fit, reference action $\mathbf a$, horizon $H$, step $\epsilon$, and thresholds $\tau_1,\tau_2,\tau_3$ fixed in advance} \\
\multicolumn{2}{@{}p{\textwidth}@{}}{\textbf{Output:} controller and  audited gradient from which it was constructed} \\[4pt]
\multicolumn{2}{@{}p{\textwidth}@{}}{\textit{Fit the surrogate}} \\
1 & fit $\widehat F_{\boldsymbol\psi}$ on the fitting transitions, minimizing $\loss_{\rm model}$ in
Equation~\eqref{eq:model-loss} \\[3pt]
\multicolumn{2}{@{}p{\textwidth}@{}}{\textit{Audit the action gradient}} \\
2 & \textbf{for} every held-out state $\mathbf x$ and every action coordinate $i$ \\
3 & \hspace{1.2em} run the solver twice from $\mathbf x$, at $\mathbf a\pm\epsilon \mathbf e_i$, to obtain $g^{\rm HF}_i$ in
Equation~\eqref{eq:paired-gradient} \\
4 & \hspace{1.2em} differentiate the objective $\widehat J_H$ predicted by $\widehat F_{\boldsymbol\psi}$, or evaluate the gradient model of Section~\ref{sec:chaotic}, \\
  & \hspace{1.2em} at the same $\mathbf x$ and $\mathbf a$ for $\widehat g_i$ in
Equation~\eqref{eq:surrogate-gradient} \\
5 & compute the pooled correlation, the sign agreement and the median
actuator correlation \\
6 & \textbf{if} all three meet their thresholds, the audited gradient is $\widehat{\mathbf g}$, which may be evaluated at further states \\
7 & \textbf{else} reject $\widehat{\mathbf g}$, and either fit a replacement surrogate and return to step 2 on new held-out states, \\
  & \hspace{1.2em} or take the measured $\mathbf g^{\rm HF}$ as the audited gradient \\[3pt]
\multicolumn{2}{@{}p{\textwidth}@{}}{\textit{Build the controller}} \\
8 & fit the direction of the extracted law $\mathbf a_{\rm law}$ to the negative audited gradient, \\
  & \hspace{1.2em} and fix its amplitude in advance or by a solver selection under a rule fixed in advance \\
9 & set the initial $\boldsymbol\phi$ so that $\pi_{\boldsymbol\phi}$ reproduces $\mathbf a_{\rm law}$, by law
projection or by adding to it a correction initialized at zero \\
10 & optimize $\boldsymbol\phi$ through $\widehat F_{\boldsymbol\psi}$, minimizing $\loss_\pi$ in
Equation~\eqref{eq:pathwise-loss}, \\
  & \hspace{1.2em} holding $\lambda_{\rm anchor}$ fixed for the anchored controller, or decaying it to zero for the released controller\\
  & \hspace{1.2em} \textbf{if} $\widehat{\mathbf g}$ was rejected, confine the action to a cone around $\mathbf a_{\rm law}$ in place of the anchor penalty \\[3pt]
\multicolumn{2}{@{}p{\textwidth}@{}}{\textit{Evaluate}} \\
11 & evaluate $\mathbf a_{\rm law}$ and $\pi_{\boldsymbol\phi}$ in the solver on held-out initial conditions \\
12 & at a new operating condition, evaluate the controller, and construct or refine a controller \\
  & \hspace{1.2em} with $\widehat{\mathbf g}$ only after an audit around the actions that controller applies passes \\
\end{tabular}
\end{algorithm}

\section{Numerical results}
\label{sec:numerical}

The four test cases assess different parts of the procedure presented in Section~\ref{sec:method}.
The rotating cylinder assesses whether a gradient that agrees with the solver where it was
measured still helps once optimization moves the policy elsewhere. The three cylinder test case assesses whether accurate prediction of the objective leads to an
accurate action gradient, and then quantifies what the anchor penalty contributes once an accurate
gradient is available. The forced turbulence test case assesses whether {a controller can be built from the measured gradient when the surrogate fails the audit, and whether controller training is repeatable} across training
seeds when the {surrogate} and its data are held fixed. Finally, the parametric chaotic equation assesses whether a controller built where the audit passes transfers to a second forcing amplitude, with no new finite differences. Together, the four test cases cover both outcomes of the audit. The surrogate passes on the rotating cylinder and on the chaotic equation, and its gradient builds the controller. On the three cylinder cluster a first surrogate fails and a replacement passes, and on forced turbulence the surrogate fails and the measured gradient builds the controller.

Every reported controller performance is obtained with a final high-fidelity evaluation on held-out initial conditions. Because the unit over which uncertainty is computed differs across the four test cases, each subsection states its own training seeds, evaluation units and limitations. %
Tables~\ref{tab:flow-overview} and~\ref{tab:protocol} summarize the configurations
and the audit outcomes, and the finite-difference trajectories behind each reported result are
counted in Section~\ref{sec:cost}, which compares the cost of the surrogate-based approach, that is, every high-fidelity simulation from data collection to the final controller within the scope stated for each test case, with model-free
training on every test case. %
More precisely, every model-based controller is compared with a model-free controller trained on the same
problem, under the same objective, action limits, and scored horizon. The three Navier Stokes cases
use proximal policy optimization (PPO), and the parametric chaotic equation uses twin delayed deep
deterministic policy gradient (TD3), the algorithm reported by its published benchmark.
The rotating cylinder and the three cylinder cluster high-fidelity solutions are obtained with HydroGym
\citep{lagemann_2025_hydrogym_platform,lagemann_2025_hydrogym}, forced two-dimensional turbulence is solved with a pseudo-spectral solver for the vorticity equation of Section~\ref{sec:governing} (more details can be found in Appendix~\ref{app:surrogate}), and the chaotic equation follows the
benchmark of \citet{botteghi2024parametric}. Appendix~\ref{app:surrogate} highlights the solver
configuration, together with the surrogate model and policy settings for each test case, and Appendix~\ref{app:scope} reports details on the evidence standard behind every reported result, namely that each controller is scored only on initial conditions excluded from its selection, and that every threshold and every selection and stopping rule is fixed before measurements are acquired. Appendix~\ref{app:per-flow-figures} shows the scores of every training seed and initial condition behind the reported metrics, Appendix~\ref{app:moved-results} collects all controller performance in one table, and Appendix~\ref{sup:baseline-settings} lists the settings of the model-free baselines.

\begin{table}[ht]
\centering
\caption{The four test cases cover boundary and volume actuations, periodic and wall bounded domains, and action spaces with dimension from one to eight. The final column states the purpose of each test.}
\label{tab:flow-overview}
\small
\begin{tabularx}{\textwidth}{>{\raggedright\arraybackslash}p{0.165\textwidth}
>{\raggedright\arraybackslash}X>{\raggedright\arraybackslash}X
>{\raggedright\arraybackslash}p{0.235\textwidth}}
\toprule
Flow and configuration & Dynamical setting & Actuation & Evaluation purpose \\
\midrule
Rotating cylinder wake, $Re=100$, one action bounded by $\pi/2$, unstructured mesh with $8{,}744$ nodes &
A circular cylinder sheds vortices from alternating sides. The resulting forces oscillate at one dominant frequency. &
The action sets the cylinder rotation rate and changes the phase of vortex shedding. &
Single action audit, comparison of anchored and unregularized policy optimization. \\
\addlinespace
Three cylinder cluster, $Re=30$, three actions bounded by $1$, unstructured mesh with $25{,}695$ nodes &
Three cylinders form a triangular cluster. Their wakes interact and can deflect to either side of the centerline but at $Re=30$ the simulated forces show no periodic oscillation. &
Each cylinder rotates independently. The three actions reshape the combined wake through the rotation rates. &
Comparison of prediction accuracy and action gradient accuracy, evaluation of anchor release. \\
\addlinespace
Forced two-dimensional turbulence, $Re=200$, four actions, $256\times256$ grid &
A periodic two-dimensional flow is sustained by a fixed sinusoidal body force and remains turbulent. &
The four actions set the amplitudes of four stationary forcing modes. &
Controller built from the measured gradient after a failed audit, and its repeatability across training seeds. \\
\addlinespace
Parametric chaotic equation, $L=22$, $64$ grid points, eight actions bounded by $1$ &
A one dimensional periodic field remains chaotic through growth, damping, and nonlinear transfer across scales. The forcing amplitude $\mu$ breaks the translational symmetry and sets the dynamical regime. &
Eight fixed Gaussian actuator profiles apply local forcing to the field. &
Transfer of the controller to a second forcing amplitude. \\
\bottomrule
\end{tabularx}
\end{table}

\begin{table}[ht]
\centering
\caption{The audit decides which gradient builds each controller. The surrogate passes in three of the five reported conditions, on the three cylinder cluster after one replacement. The measured gradient builds the controller on forced turbulence, and the controller built at $\mu=0.225$ is transferred to $\mu=0.121$. In all five conditions the surrogate-based controller improves on the model-free baseline or is not distinguished from it. Audit configuration and final high-fidelity outcome for the five reported conditions. On the three flow test cases the values are reductions of the objective in \%, and higher is better. On the chaotic equation they are accumulated state costs, and lower is better. Each uncertainty is the sample standard deviation over the units stated in the corresponding subsection. {The audit criteria are the pooled correlation, the sign agreement and the median actuator correlation, in this order.}
}
\label{tab:protocol}
\footnotesize
\begin{tabularx}{\textwidth}{>{\raggedright\arraybackslash}p{0.135\textwidth}
>{\raggedright\arraybackslash}p{0.145\textwidth}>{\raggedright\arraybackslash}p{0.145\textwidth}
>{\raggedright\arraybackslash}p{0.145\textwidth}>{\raggedright\arraybackslash}p{0.125\textwidth}
>{\raggedright\arraybackslash}X}
\toprule
 & Rotating cylinder & Three cylinder cluster & Forced turbulence & Chaotic $\mu=0.225$ & Chaotic $\mu=0.121$, transferred \\
\midrule
Surrogate & latent field model & force model, then latent field model & Fourier neural operator & gradient model & gradient model from $\mu=0.225$, not refitted \\
\addlinespace
Audit result & pass, {$0.996$, $1.00$, one actuator} & {force model} fail{s at} {$0.459$, $0.61$, $0.51$}, {latent field model passes} at {$0.938$, $0.97$, $0.91$, then $0.936$, $0.94$, $0.64$} & {fail, $0.53$, $0.55$, $-0.34$} & pass, {$0.979$, $0.95$, $0.89$} & {inherited from $\mu=0.225$} \\
\addlinespace
Gradient building the controller & surrogate gradient & gradient of the latent field model, after the force model is rejected & measured gradient & gradient model & none at $\mu=0.121$, the controller is transferred from $\mu=0.225$ \\
\addlinespace
Reference action, horizon & zero, $H=25$ & zero, $H=1$ & zero, $H=20$ & action of the base controller, $H=50$ & as at $\mu=0.225$ \\
\addlinespace
Audited states & eight held-out states spread evenly over one vortex shedding cycle & 55 held-out states, 38 of them also used for fitting, then 16 new held-out states & 20 uncontrolled states & eight held-out states from one trajectory & as at $\mu=0.225$ \\
\addlinespace
Controller &  cosine of the shedding phase minus a fitted phase (Equation~\eqref{eq:distilled-law}) & linear map from {the nine measured scalars} to rotation rates & neural state feedback within $5^\circ$ of the measured descent direction & {neural policy} plus a correction from the gradient model & {controller from $\mu=0.225$, with sensor noise filtered by a linear observer} \\
\addlinespace
Baseline, our approach & PPO $0.21\pm3.54$, {ours} $3.83\pm0.14$ & PPO $17.57\pm0.30$, ours $17.10\pm0.26$ & PPO {$30.63\pm21.43$}, ours $82.01\pm0.19$ & TD3 $409.56$, ours $361.58\pm8.31$ & TD3 $71.93$, ours $74.92\pm4.01$ \\
\addlinespace
Main finding & {controller has  higher mean reduction of objective function than baseline} & prediction accuracy and gradient accuracy differ,  final comparison detects no difference & {controller built from measured gradient} improves on {four seed} baseline & {no detected difference from TD3, at 108 times fewer solver interactions} & {noninferior to TD3 once sensor noise is filtered, at 156 times fewer solver interactions} \\
\bottomrule
\end{tabularx}
\end{table}

\subsection{Rotating cylinder wake}
\label{sec:latentfail}

The rotating cylinder wake establishes that {a surrogate can pass the audit and that its gradient then yields an effective controller}, while optimizing a policy through the same surrogate without the anchor penalty {results in a poor controller. The surrogate meets both acceptance criteria that apply to a single actuator at $Re=100$, so its descent directions, which agree with those measured with the high-fidelity solver, are used to fit the extracted law. %
Finally, the same law is applied at three higher Reynolds numbers not used to fit either the surrogate or the policy. Figure~\ref{fig:cylinder-solution} shows the flow with and without control and compares the controllers.

The cylinder experiences a drag force along the flow direction and, because it sheds vortices alternately from its two sides, a lift force that oscillates at the shedding frequency. The drag and lift coefficients are written as $C_D$ and $C_L$. The objective and the reward used by model-free training at step $t$ are
\begin{equation*}
    J(\pi)=\overline{C_D}+0.2\,\overline{|C_L|},
    \qquad
    r_t=-\Delta t\,(C_{D,t}+0.2|C_{L,t}|),
\end{equation*}
where the overbar denotes an average over the full evaluation window and $\Delta t$ is
the control interval. The lift term has a weight equal to $0.2$ because a steady rotation can lower the drag while leaving a large {lift} force on the cylinder, and an objective built from
drag alone would reward that solution. Appendix~\ref{app:cylinder} gives the solver configuration, and Appendix~\ref{sup:phase-frame} reports the computation of the shedding phase.

Since the wake oscillates at one dominant frequency, the phase
$\varphi(t)$ of that oscillation summarizes the state of the flow, and
the measured action gradient varies with it as a single harmonic, so
its average over the sampled states vanishes. Consequently, a constant
rotation gives no first-order reduction of the objective, and the
controller must depend on the phase. Fitting a single harmonic of the
phase to the descent directions of the audited latent surrogate, at 18 states visited by the controllers trained through it, gives the law
\begin{equation}
    a(t)=\alpha\,\cos\!\big(\varphi(t)-\varphi_{\rm law}\big),
    \qquad
    \varphi_{\rm law}=\varphi_0+180^\circ ,
    \label{eq:distilled-law}
\end{equation}
where $\alpha$ bounds the commanded rotation and $\varphi_0$ is the phase of the harmonic fitted to
the surrogate gradient. The offset of $180^\circ$ turns that gradient into a descent direction. The law has two coefficients, the amplitude $\alpha$, fixed in advance at $0.2$, and the phase $\varphi_{\rm law}$, which is determined by the least-squares fit of Section~\ref{sec:training}, thus obtaining $\varphi_{\rm law}=164^\circ$. {The audit shows that this surrogate gradient agrees with the measured one, with correlation $0.996$ at the eight audited states, and a harmonic fitted to the measured gradient at the same 18 states gives $163^\circ$.} %

The law built from the audited gradient has a higher mean reduction than matched model-free training, as Figure~\ref{fig:cylinder-solution}{e} shows. It reduces the objective by $3.83\pm0.14\%$ over eight unseen initial conditions. Each test performed compares a controlled run with an uncontrolled one started from the same initial state, which gives one reduction of the objective per initial condition. The mean of these values is $3.83$, and $0.14$ is their standard deviation. Matched model-free training reaches $0.21\pm3.54\%$ over 12 paired rollouts. These results are obtained by three independently trained seeds, each scored on four unseen initial conditions after $32{,}768$ solver interactions, and the three seeds quantify how much model-free training varies across runs on this test case. The standard deviation of the 12 {reductions} exceeds their mean, so model-free training provides no reliable reduction here. A Welch test, a two-sample test that does not assume equal variances, compares the eight objective reductions, one per initial condition, with the three PPO seed means, one per training seed. It gives a difference of $3.62$ percentage points with a $95\%$ interval from $-6.10$ to $13.35$, so the difference is not statistically significant. %
Model-free training had not converged within its budget, since its mean over seeds moves from $1.82\%$ to $-0.64\%$ and then to $0.21\%$ between $8{,}192$, $16{,}384$ and $32{,}768$ interactions. Appendix~\ref{sup:ppo-curve} reports the score of each seed at three points during training, and the value obtained by choosing each seed's best checkpoint on the evaluation data.

An accurate gradient on its own does not produce a working controller. Indeed, pathwise optimization through
the audited surrogate, run without the anchor penalty and from an initialization not derived from the law, produces $-0.52\pm{0.25}\%$ over three training seeds, where the uncertainty is the sample standard deviation across the seeds. Each seed value is the best of eight training iterations, each scored by one rollout against the mean uncontrolled objective, and the final iterations give $-1.25\pm0.52\%$. The negative value means that the objective ends higher than with no control. The surrogate is accurate where it was audited, since its action gradient over the 25-step horizon agrees with the paired high-fidelity measurement at correlation
$0.996$ and a phase error of $0.8^\circ$. The failure comes from the policies that optimization reaches. Training through the surrogate moves toward policies that apply their largest rotation at shedding phases between $77^\circ$ and $122^\circ$, while the measured gradient places them at phases from $124^\circ$ to $163^\circ$. Initializing by law projection and holding the
anchor before reducing it to zero keeps the policy in the supported region. %
The released
controller reduces the objective by $3.33\pm0.15\%$, and similar results hold for all four training seeds.
That uncertainty is the sample standard deviation across 32 paired rollouts, four training runs
scored on the same eight initial conditions, so the 32 rollouts are not independent. On this test, pathwise training does not improve on the extracted law, which reduces the objective by $3.83\%$. The role of the neural policy on this test case is to show that optimization started from the law and held near it stays in the supported phase region, while optimization started elsewhere does not.

\begin{figure}[ht]
\centering
\includegraphics[width=\textwidth]{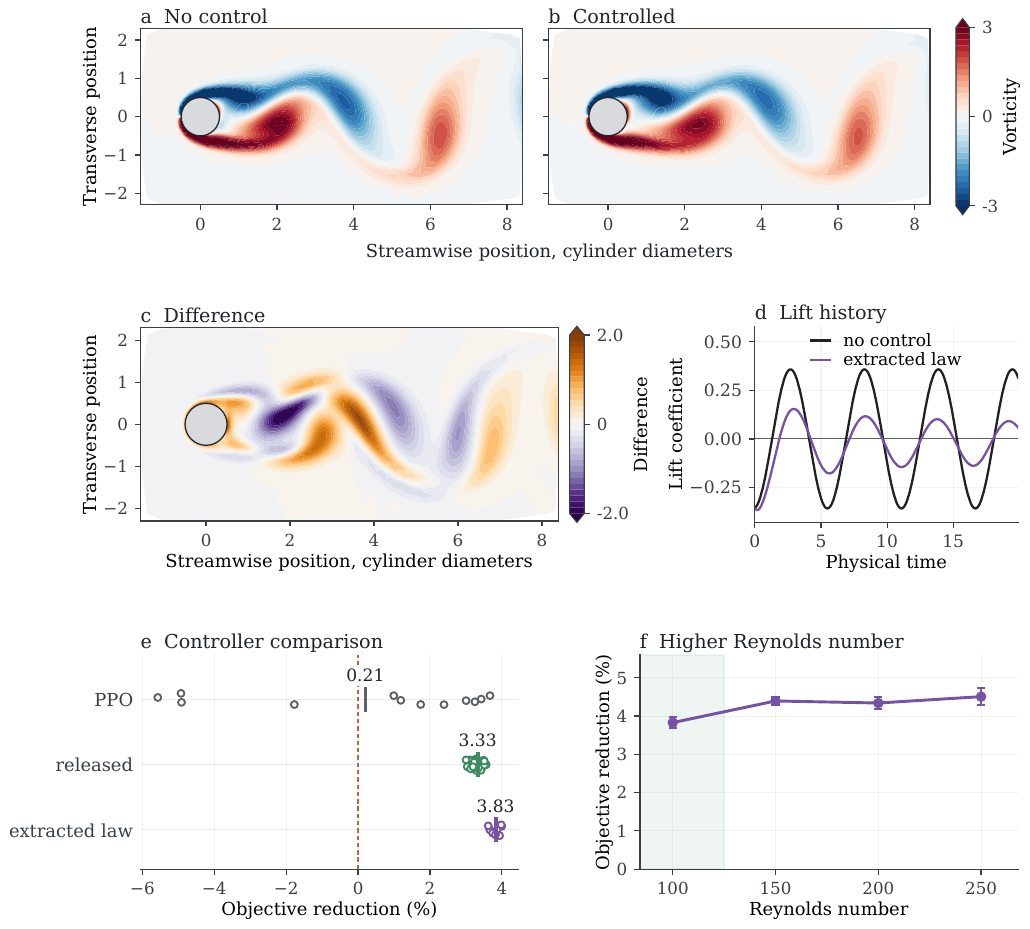}
\caption{The extracted law weakens the alternating wake vortices and the oscillation of the lift force, and the same law keeps its effect at higher $Re$. Panel a shows the uncontrolled vorticity field and panel b the controlled field from the same initial state. Panel c shows their difference so that the regions of the wake changed by the control are visible. Panel d compares the lift coefficient over time in the two cases. Panel e reports the reduction of the objective against a paired uncontrolled rollout from the same initial
state, for matched model-free PPO over 12 paired rollouts, the released controller over 32
paired rollouts, and the extracted law over eight initial conditions. The dashed line marks zero reduction, and a point {to its right} is a controller that improves the objective. Panel {f} applies the {extracted law obtained at $Re=100$} at Reynolds numbers 150, 200 and 250 with no refitting, and the shaded band marks
the condition at which it was fitted.}
\label{fig:cylinder-solution}
\end{figure}

The {extracted} law transfers to higher Reynolds number without being refitted. {Indeed, the extracted law obtained at $Re=100$ is applied unchanged at Reynolds numbers $150$, $200$ and $250$, without new finite differences, as Figure~\ref{fig:cylinder-solution}{f} shows. A neural policy is not used in this test.} {The law} reduces the objective by $4.39\pm0.10$, $4.34\pm0.15$ and $4.51\pm0.22\%$ at the three $Re$, over eight initial conditions each and scored against an uncontrolled baseline{, where the uncertainty is the sample standard deviation across the eight conditions}. %
{The phase estimate uses the shedding period of $Re=100$, and replacing it by the period measured at $Re=200$ improves the performance by $0.16$ percentage points, as Appendix~\ref{app:cylinder} reports.}

\subsection{Three cylinder cluster}
\label{sec:repair}

The three cylinder cluster establishes that a surrogate can predict the objective almost exactly and
still misestimate the action gradient, so that the audit rejects the surrogate. It then assesses what the anchor penalty contributes once an accurate gradient is available. Because the simulated wake shows no periodic oscillation at $Re=30$, the measured gradient varies little from state to state. Consequently, the case isolates the effect of the anchor penalty and assesses state-dependent feedback only weakly.

The three cylinders stand in a triangular arrangement and each one rotates independently, so the
action has three components bounded by one. Writing $C_{D,j}$ and $C_{L,j}$ for the drag and {lift} coefficients of cylinder $j$, the objective is
\begin{equation}
    J_{\rm cluster}
    =\sum_{j=1}^{3}C_{D,j}
    +0.2\left|\sum_{j=1}^{3}C_{L,j}\right|,
    \label{eq:pinball-objective}
\end{equation}
for which lower values are better. The high-fidelity evaluation applies no penalty on the action, so
every reduction below is a reduction of the flow objective alone. Appendix~\ref{app:pinball} reports the solver configuration.%

Two surrogates were fitted for this flow, and they differ in the quantities each one observes. The force model takes as input nine scalars, the lift, drag and actuator value of each cylinder, while the latent field model uses the velocity, pressure and vorticity fields on the mesh through an encoder. The
force model predicts the objective almost exactly, at correlation $0.9996$ on held-out data. This is expected, because
Equation~\eqref{eq:pinball-objective} builds the objective from those same forces, but its action gradient is far less accurate. Indeed, measured against the solver over 72 held-out comparisons, 24 states with three actions each, it correlates at $0.459$ and misses the acceptance criteria, so no controller is built
from it. The nine scalars do not contain the state of the controlled wake, which determines how the flow responds to a rotation. The latent field model takes that state as input and reaches $0.938$ on 72 held-out {gradient labels} from its own data set, and $0.936$ on a second set of 16 new held-out states{, and passes the audit on both}, as the replacement rule of Algorithm~\ref{alg:procedure} requires. We remark that the audit decision matters for the controller. Fitting the same law to the action gradient of the rejected force model, on the same 72 states and with the same gain, gives a controller that raises the objective by $18.04\%$ on the same eight conditions. The paired difference from the law fitted to the accepted surrogate is $34.08$ percentage points, as Appendix~\ref{app:pinball} reports.

The extracted law here is a linear map from the nine measured scalars to the three rotation rates, and its coefficients are fitted to the descent directions of the latent field model, which passed the audit. Because the gradient fixes {only} the direction of the law and leaves its size undetermined, the gain is selected in the solver from a grid fixed in advance, on three initial conditions reserved for that selection, as Appendix~\ref{app:pinball} describes. With the selected gain, the control reduces the objective by $16.04\pm0.01\%$ on eight held-out initial conditions, where the uncertainty is the sample
standard deviation across those conditions.

The anchored controller reaches an objective reduction {that the available seeds do not distinguish} from matched model-free training, at 24 times fewer solver interactions}. Holding the anchor weight fixed reduces the objective by $17.10\pm0.26\%$ across four training seeds, with all 32 paired rollouts positive. {Matched} model-free training reduces it by $17.57\pm0.30\%$ across three training seeds, and each uncertainty is the sample standard deviation across seed means. A Welch test on the seven seed means returns $t=-2.14$ on $4.1$ degrees of freedom at $p=0.10$. Consequently, these data detect no difference between the two, and seven seeds give the test too little power to establish that they perform equally. The complete surrogate-based approach uses {$4{,}930$} solver interactions against $120{,}320$ for each model-free training seed. Therefore, the two reach comparable reductions at costs that differ by a factor of {24}.

\begin{figure}[ht]
\centering
\includegraphics[width=\textwidth]{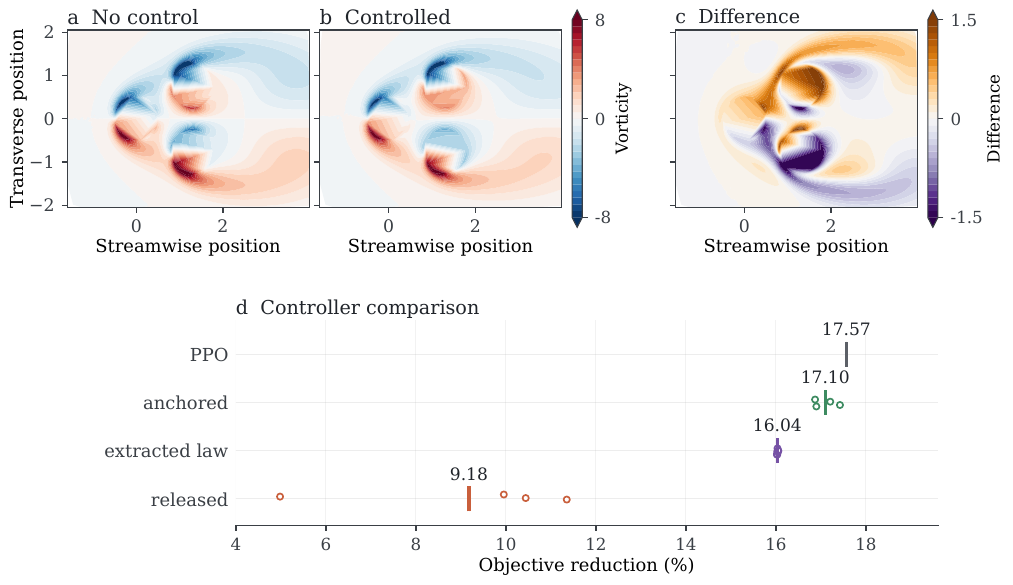}
\caption{{The extracted law reduces the objective almost as much as model-free PPO, the anchored controller built from it improves this reduction, and the released controller loses almost half of it.} Panel a shows the uncontrolled vorticity field, panel b the {field controlled by the extracted law,}
from the same initial state, and panel c their difference on a separate
scale. Panel d reports the reduction of the objective against paired uncontrolled rollouts from the
same initial states, for {matched} model-free PPO over three training seeds, the anchored controller over
four training seeds, the extracted law over eight initial conditions, and the released controller over four
training seeds.}
\label{fig:pinball-solution}
\end{figure}

Releasing the anchor costs about half of the anchored reduction. Indeed, reducing {the anchor weight} to zero decreases it to $9.18\pm2.86\%$ on the same four training seeds, and widens the spread across those seeds by roughly an order of magnitude. The released controllers leave a larger net lift force on the cluster, which Equation~\eqref{eq:pinball-objective} penalizes. The absolute value of the total lift coefficient of the three cylinders, averaged over time, is $0.419$ against $0.137$ for the anchored controllers, while the average size of the action is almost the same, $1.36$ against $1.34$, as Appendix~\ref{app:pinball} reports. Figure~\ref{fig:pinball-solution} shows the controlled flow and compares the four controllers.

\subsection{Forced two-dimensional turbulence}
\label{sec:kolmogorov}

The forced two-dimensional turbulence test case gives the largest improvement over model-free training, and the controller holds its action near the measured descent direction by construction instead of {using} a penalty. Its architecture admits only actions within five degrees of the mean measured descent direction. The anchor weight is accordingly $\lambda_{\rm anchor}=0$.
Appendix~\ref{app:turbulence} reports a separate controller trained with the anchor penalty on this test case, together with its released counterpart.

The controller minimizes enstrophy, the spatial average of the squared vorticity
\begin{equation*}
    \Phi=\langle\omega^2\rangle,
\end{equation*}
where the angle brackets denote the average over the periodic domain. Enstrophy measures the intensity of the small-scale vorticity. It is the natural target here because in two dimensions energy moves toward large scales, while enstrophy moves toward small scales, where viscosity dissipates it \citep{boffetta2017two}.
The four actions set the amplitudes of the forcing modes of
Equation~\eqref{eq:kolmogorov-force}, and Appendix~\ref{app:surrogate} gives the grid and the
observation field. The reported reduction is the paired gain of Equation~\eqref{eq:paired-gain}, with the enstrophy averaged over the final 10 control steps and compared with the uncontrolled scenario from the same initial condition.

On this flow the surrogate, a Fourier neural operator that advances the vorticity field, does not meet the acceptance criteria of Section~\ref{sec:audit}. Indeed, over 20 states sampled from the uncontrolled flow at zero reference action, the pooled correlation is $0.53$, the sign agreement is $0.55$, and the median actuator correlation is $-0.34$, all below their thresholds. More precisely, the surrogate recovers the dominant and nearly state-independent part of the gradient field, which is the direction the controller uses. However, it reproduces the state-to-state variation poorly: the correlations computed across states, one actuator at a time, are all negative, and the pooled value of $0.53$ is mainly related to the common part. {Each of these four correlations compares the surrogate and measured values of one gradient component across the 20 states, and they range from $-0.65$ to $-0.02$. The pooled correlation combines all four components, and it is positive because in both gradients the first component is much larger than the others and has the same sign at every state.} %
The per-state $\cos\theta$ of Section~\ref{sec:threshold} between the surrogate and high-fidelity gradients is positive at all 20 states, with median $0.86$. Consequently, at every audited state the surrogate gradient gives a descent direction of the high-fidelity objective, although its variation across states is wrong.

Since the surrogate fails the audit, this case follows the measured-gradient branch of Algorithm~\ref{alg:procedure} (step 7). The extracted law is set by the measured gradient, and the surrogate is used only to optimize a small state-dependent residual, that is, a correction added to the action of the extracted law. {This correction is produced by a neural policy, a small network that reads the lowest wavenumber Fourier coefficients of the vorticity field and is trained on trajectories predicted by the surrogate, by differentiating the predicted objective as in Section~\ref{sec:training}.} On this test case the extracted law is a constant action along the negative direction of the {forcing mode with amplitude $a_4$}, a direction about $3^\circ$ from the mean measured descent direction, and the residual is confined to a cone of five degrees around that action. {The} controller design, {the} training seeds and {the acceptance rule for trained controllers} {are} fixed before any controller {is} trained. Each trained controller is accepted only if two conditions hold, as Appendix~\ref{app:turbulence} describes. The median cosine between its action and the measured descent direction must be at least $0.70$, and its mean action must lie within $15^\circ$ of the action of the extracted law. We call this condition the controller check, to distinguish it from the acceptance criteria of the audit. Five degrees was not selected on controller performance. All eight controllers pass the check, with median cosine between $0.989$ and $0.990$.

\begin{figure}[ht]
\centering
\includegraphics[width=\textwidth]{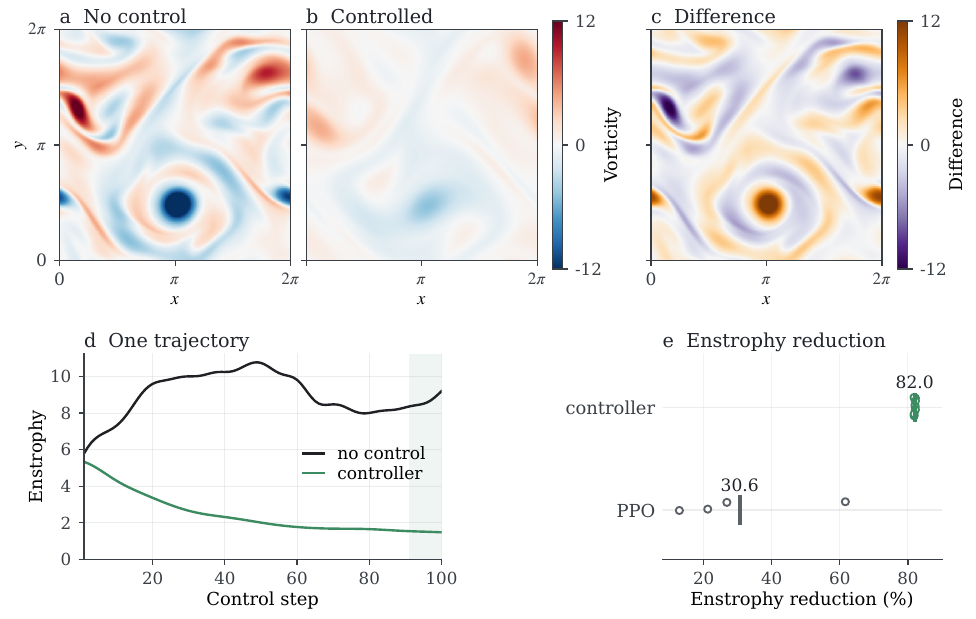}
\caption{The controller {weakens} the vortices and reduces enstrophy for every training seed. Panel
a shows the uncontrolled final vorticity field, panel b the controlled field from the same initial
state, and panel c their difference on a separate scale. Panel d shows enstrophy over time for one initial condition, with and without control, from the first  step onward. The shaded interval marks the final 10 control steps, over which enstrophy is averaged to evaluate a controller. Over this interval enstrophy is $82.5\%$ lower with the proposed controller. This value comes from a single trajectory, so it differs slightly from the $82.01\%$ averaged over all seeds and evaluation conditions in the text. Panel e shows the held-out enstrophy reduction,
whose upper row holds one value for each of the eight controller training seeds and whose lower row
holds one value for each of the four PPO training seeds, each the mean over the same eight evaluation conditions.}
\label{fig:kolmogorov-solution}
\end{figure}

{The enstrophy reduction of the resulting controller is about} {$2.7$} {times that of model-free training. Across the eight training seeds it reduces enstrophy by $82.01\pm0.19\%$, where $0.19$ is the sample standard deviation of the eight seed means, each averaged over eight held-out initial conditions}. {Matched PPO, trained with four independent seeds and evaluated on the same eight held-out initial conditions, reduces enstrophy by $30.63\pm21.43\%$, where $21.43$ is the sample standard deviation of the four seed means. Its result varies strongly from one training run to another, between $12.88$ and $61.61\%$. The advantage of the controller, obtained through the proposed approach, is $51.38$ percentage points, with a Welch $95\%$ confidence interval from $17.28$ to $85.48$ percentage points computed over the training seeds of both methods.} The performance criterion fixed before the runs consists of three conditions: at least seven of the eight controllers pass the controller check, the solver diverges in no evaluation, and the lower end of a bootstrap interval on the advantage over PPO, obtained by resampling training seeds, is positive. All three are met, the bootstrap condition was fixed against the single PPO seed available at that time, and it also holds with all four PPO seeds, for which the interval runs from $30.48$ to $65.67$ percentage points. Figure~\ref{fig:kolmogorov-solution} shows the controlled flow and the reduction for every seed.

The eight seeds show that controller training is repeatable. They share one
surrogate, one set of audited states and one set of finite-difference labels, so their agreement
does not indicate how much the result would change if the surrogate is refitted.

\subsection{Parametric chaotic equation}
\label{sec:chaotic}

The parametric chaotic equation establishes {that a controller built where the audit passes transfers to a second flow condition, and} that an audit pass is local to the actions at which it
was measured{, as Appendix~\ref{app:chaotic} shows}. {Two values of the forcing amplitude are considered, $\mu=0.225$ and $\mu=0.121$.} At the higher {value} the audit passes at a fixed reference action and the
resulting controller matches the performance of the algorithm used in the published reference benchmark \citep{botteghi2024parametric}. At the lower
value the controller is transferred without new finite differences or retraining, and it is noninferior to the benchmark algorithm once the noise is filtered out of the sensor readings it receives. 

In \citet{botteghi2024parametric}, the forcing
amplitude $\mu$ is supplied to the controller. More precisely, writing $u_j$ for the field at the 64 grid points and $a_k$ for
the eight actuator amplitudes, the state cost and the action cost at each control step are defined as
\begin{equation*}
    c_1(t)=\frac{1}{64}\sum_{j=1}^{64}u_j(t)^2,
    \qquad
    c_2(t)=\frac{1}{8}\sum_{k=1}^{8}a_k(t)^2,
\end{equation*}
and the reward is $-c_1-0.1c_2$. Performance is reported as {the state cost $c_1$ accumulated over an episode, which we call the accumulated state cost, and lower values are better.} Appendix~\ref{app:surrogate} gives the discretization and 
sensor configuration.

{On this test case the controller is built at $\mu=0.225$ in two stages, each following Algorithm~\ref{alg:procedure}. In the first stage, a transition surrogate, which passes its own audit as Appendix~\ref{app:surrogate} reports, is used to fit a linear feedback law from the observation to the eight actions and to train anchored neural policies near that law, as in Section~\ref{sec:training}. We call these policies the base policies, and they also supply the actions that follow the perturbed first action in the audit of Section~\ref{sec:audit}. In the second stage, the controller adds to the base action a correction of bounded size along the negative action gradient of the state cost accumulated over 50 control steps. Differentiating the transition surrogate over 50 steps would compound its error, so this gradient is predicted directly by a separate network, called the {gradient model}. The gradient model takes the eight sensor values and $\mu$ as input, it is fitted to the finite differences of Equation~\eqref{eq:paired-gradient} measured along the trajectories of the base policies, with all states of one trajectory on the same side of the division between fitting and held-out states, and it must pass the audit before it is used. The corrected action then passes through fixed transformations, the last of which moves each action component a fraction $r$ of the way toward $0.999$ times its bound while keeping its sign, where the matched TD3 controllers also operate. We call $r$ the strength of this transformation, and for this controller $r=0.90$. Each of these choices was made by a rule fixed in advance, on initial fields disjoint from the evaluation fields.}

\begin{figure}[ht]
\centering
\includegraphics[width=\textwidth]{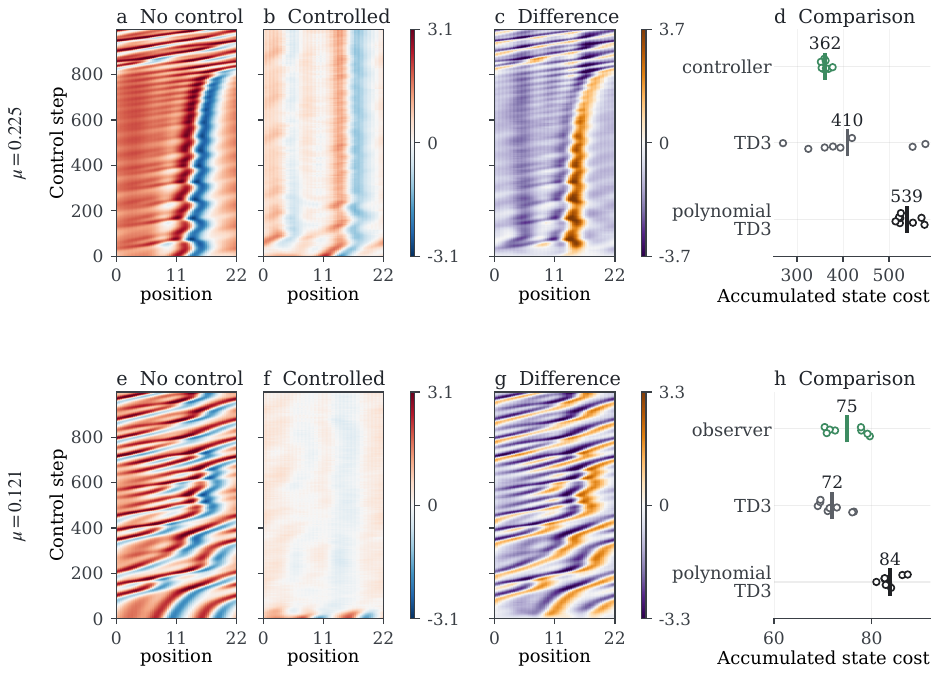}
\caption{{The surrogate-based controller weakens the field at both forcing amplitudes, at a cost close to that of matched TD3 and below that of matched polynomial TD3.} Panels a to c show the uncontrolled field, controlled field, and their difference at $\mu=0.225$. Panels e to g show the corresponding fields for the {controller with observer} at $\mu=0.121$. Each controlled field and its uncontrolled pair start from the same initial field. Panels d and h compare the controller with {matched} TD3 and {matched} polynomial TD3 on {the eight held-out initial fields}. {The row labeled controller in panel d is the surrogate-based controller at $\mu=0.225$, and the row labeled observer in panel h is the controller with observer at $\mu=0.121$.} Lower {accumulated} state cost is better.}
\label{fig:ks-solution}
\end{figure}

{At $\mu=0.225$ the gradient model passes the audit, with pooled correlation $0.979$ on eight held-out states from one trajectory, and the resulting controller is not distinguished from TD3, the algorithm of the benchmark. Over eight held-out initial fields it reaches an accumulated state cost of $361.58\pm8.31$, where the uncertainty is the sample standard deviation across the fields, against $409.56$ for matched TD3, and the $95\%$ confidence interval of their paired difference includes zero. Its cost is also lower than that of matched polynomial TD3, the variant of TD3 with a sparse polynomial policy proposed by \citet{botteghi2024parametric}, on all eight fields, with a mean of $538.71$ for polynomial TD3. Table~\ref{tab:ks-scores} in Appendix~\ref{app:chaotic} gives the paired comparisons, and Figure~\ref{fig:ks-solution} shows the controlled fields at both forcing amplitudes.}

{The controller built at $\mu=0.225$ is then transferred to $\mu=0.121$ without new finite differences or retraining. Only the strength of the last transformation was chosen again at $\mu=0.121$, on initial fields disjoint from the evaluation fields, and it became $r=0.999$. The protocol fixed in advance for this transfer had set $r=0.90$, so this choice departs from it. Applied to the raw sensor readings, the transferred controller reaches an accumulated state cost of $134.98$, against $71.93$ for matched TD3, and the gap comes mainly from sensor noise. The benchmark adds sensor noise of standard deviation $0.25$, and under control the root mean square of the noise-free sensor signal is $0.959$ times this standard deviation at $\mu=0.121$, against $2.051$ times at $\mu=0.225$. Consequently, at the lower amplitude the signal is no larger than the noise, and on four further fields removing the noise lowers the cost of the transferred controller from $136.19$ to $65.14$.}

{Filtering the noise out of the sensor readings makes the transferred controller noninferior to matched TD3. A linear observer predicts the clean sensor values one step ahead from its previous estimate and the applied action, and in each update it combines this prediction with the noisy measurement, which receives a weight of one tenth. The observer is fitted on noise-free transitions and then kept fixed, and the controller receives its estimates in place of the raw sensor values without being retrained. Over eight held-out initial fields the controller with observer reaches an accumulated state cost of $74.92\pm4.01$, against $71.93$ for matched TD3. Their paired difference is $+2.99$, with a $95\%$ confidence interval from $-1.09$ to $+7.08$, whose upper end lies inside the noninferiority margin of $7.19$, that is, $10\%$ of the TD3 cost, fixed before the comparison. Its cost is also $8.83$ lower than that of matched polynomial TD3, with a $95\%$ confidence interval from $5.91$ to $11.74$. Appendices~\ref{sup:ks-protocol} and~\ref{app:moved-results} report the observer fit and the selection protocol.}

This test case shows that a controller built where the audit passes transfers to a second forcing amplitude without new finite differences or retraining, and that it is noninferior to matched TD3  once its sensor input is filtered.

\subsection{High-fidelity simulation cost}
\label{sec:cost}

Including the cost of the audit, the surrogate-based approach uses fewer high-fidelity solver interactions than model-free training on every test case, although on forced turbulence only part of its cost can be counted. For $S$ sampled states, $m$ action components and a horizon $H$, the central differences of Equation~\eqref{eq:paired-gradient} require $2SmH$ solver interactions, in addition to the unperturbed trajectories from the sampled states.

On the rotating cylinder the proposed {approach} uses between $3.0$ and $4.4$ times fewer solver interactions than one model-free training run. The values are related to how the stored data sets are counted. Several of them hold up to four identical copies of the state from which a finite difference starts. Consequently, the finite differences behind the reported result cost $2{,}766$ solver interactions when each distinct state is counted once and $6{,}264$ when every stored copy is counted. Finite differences account for 38 to $58\%$ of the surrogate-based total under those two counting rules.

On the remaining three test cases the approach is cheaper by larger factors. It uses 24 times fewer interactions on the three cylinder cluster and 108 times fewer on the chaotic equation at $\mu=0.225$. On the three cylinder cluster this count leaves out the 4,680 solver interactions that selected the gain of the law, and including them gives a factor of 12. Both factors are measured against
one model-free training run, and the chaotic baseline is the published figure of $9{,}990{,}000$
solver interactions. At $\mu=0.121$ the controller is transferred from $\mu=0.225${. The}
accounting covers {the runs that chose the strength $r$ {of the last action transformation, defined in Section~\ref{sec:chaotic},} at $\mu=0.121$ and the fitting, selection and held-out evaluation of the observer}, and the proposed approach uses {156} times fewer interactions {than the published TD3 training run}. {For forced turbulence the accounting covers the audit and the controller but not the high-fidelity transitions used to fit the surrogate, which were not kept. On this partial count the approach uses 6.9 times fewer interactions than one PPO training run.}

On the three flow test cases the model-free reference is a training run carried out for this study under the same protocol, so the factors of $3$ to $24$ compare the two approaches directly. On the chaotic equation the reference is the training budget fixed by the published benchmark, so the factors of $108$ and $156$ also reflect the length of that training run. Appendix~\ref{app:cost-repro} reports the stored counts and the scope of each comparison.

\section{Conclusions}
\label{sec:discussion}

In this work, we proposed a pathwise procedure that audits the surrogate action gradient against the high-fidelity solver and anchors the policy to a law fitted from the audited gradients. Four test cases assess it, and the following conclusions are supported.

First, prediction accuracy does not imply gradient accuracy. %
Second, a passing audit does not produce an effective controller, on its own, or predict its performance. %
Third, a pass is local to the actions at which the gradient is measured. %
Fourth, the surrogate-based approach is cheaper than model-free training, and in no test case is it found to perform worse. With the cost of the audit included, it requires fewer high-fidelity solver interactions on every test case.%

However, several challenges remain. %
The acceptance statistics are uninformative when a single direction dominates the gradient field, as in forced turbulence. The test cases are two-dimensional flows at low Reynolds number and a one-dimensional chaotic equation, and extending the procedure to three-dimensional flows is the natural next step. Future work should focus on addressing these limitations and on developing an audit that remains valid as the controller changes along the optimization trajectory, a rule for deciding in advance whether to release the anchor, and acceptance statistics suited to nearly state-independent gradient fields.

\subsubsection*{Reproducibility Statement}

Appendix~\ref{app:scope}
states the rules used to classify the evidence, and the appendices that follow it report the protocol, fixed thresholds,
selection rules and stopping rules for each test case.
Appendix~\ref{app:cost-repro} gives the simulation cost accounting.

The code, the protocols fixed before each run, the environment specification, the run configurations, the stored result files, and a list that maps each reported value to its result file, with a checksum of every file, are available at \url{https://anonymous.4open.science/r/joint_mbrl-D1EC/}.

\subsubsection*{Acknowledgments}

The authors acknowledge support from the National Science Foundation AI Institute in Dynamic
Systems, grant number 2624398.

\bibliographystyle{tmlr}
\bibliography{references}

\begin{thebibliography}{38}
\providecommand{\natexlab}[1]{#1}
\providecommand{\url}[1]{\texttt{#1}}
\expandafter\ifx\csname urlstyle\endcsname\relax
  \providecommand{\doi}[1]{doi: #1}\else
  \providecommand{\doi}{doi: \begingroup \urlstyle{rm}\Url}\fi

\bibitem[Amos et~al.(2021)Amos, Stanton, Yarats, and Wilson]{amos2021modelbasedsvg}
Brandon Amos, Samuel Stanton, Denis Yarats, and Andrew~Gordon Wilson.
\newblock On the model-based stochastic value gradient for continuous reinforcement learning.
\newblock In \emph{Proceedings of the 3rd Conference on Learning for Dynamics and Control}, volume 144 of \emph{Proceedings of Machine Learning Research}, pp.\  6--20. PMLR, 2021.
\newblock URL \url{https://proceedings.mlr.press/v144/amos21a.html}.

\bibitem[Beck(2017)]{beck2017first}
Amir Beck.
\newblock \emph{First-Order Methods in Optimization}.
\newblock MOS-SIAM Series on Optimization. Society for Industrial and Applied Mathematics, Philadelphia, PA, 2017.
\newblock \doi{10.1137/1.9781611974997}.

\bibitem[Bergmann et~al.(2005)Bergmann, Cordier, and Brancher]{bergmann_2005_pod_rotary}
Michel Bergmann, Laurent Cordier, and Jean-Pierre Brancher.
\newblock Optimal rotary control of the cylinder wake using proper orthogonal decomposition reduced-order model.
\newblock \emph{Physics of Fluids}, 17\penalty0 (9):\penalty0 097101, 2005.
\newblock \doi{10.1063/1.2033624}.

\bibitem[Boffetta \& Ecke(2012)Boffetta and Ecke]{boffetta2017two}
Guido Boffetta and Robert~E. Ecke.
\newblock Two-dimensional turbulence.
\newblock \emph{Annual Review of Fluid Mechanics}, 44:\penalty0 427--451, 2012.
\newblock \doi{10.1146/annurev-fluid-120710-101240}.

\bibitem[Botteghi \& Fasel(2024)Botteghi and Fasel]{botteghi2024parametric}
Nicol{\`o} Botteghi and Urban Fasel.
\newblock Parametric {PDE} control with deep reinforcement learning and differentiable {L0}-sparse polynomial policies.
\newblock \emph{arXiv preprint arXiv:2403.15267}, 2024.
\newblock URL \url{https://arxiv.org/abs/2403.15267}.

\bibitem[Brunton \& Noack(2015)Brunton and Noack]{brunton2015closed}
Steven~L. Brunton and Bernd~R. Noack.
\newblock Closed-loop turbulence control: Progress and challenges.
\newblock \emph{Applied Mechanics Reviews}, 67\penalty0 (5):\penalty0 050801, 2015.
\newblock \doi{10.1115/1.4031175}.

\bibitem[Chua et~al.(2018)Chua, Calandra, McAllister, and Levine]{chua2018deep}
Kurtland Chua, Roberto Calandra, Rowan McAllister, and Sergey Levine.
\newblock Deep reinforcement learning in a handful of trials using probabilistic dynamics models.
\newblock In \emph{Advances in Neural Information Processing Systems (NeurIPS)}, volume~31, 2018.
\newblock URL \url{https://papers.nips.cc/paper_files/paper/2018/hash/3de568f8597b94bda53149c7d7f5958c-Abstract.html}.

\bibitem[Czarnecki et~al.(2017)Czarnecki, Osindero, Jaderberg, {\'S}wirszcz, and Pascanu]{czarnecki2017sobolev}
Wojciech~Marian Czarnecki, Simon Osindero, Max Jaderberg, Grzegorz {\'S}wirszcz, and Razvan Pascanu.
\newblock Sobolev training for neural networks.
\newblock In \emph{Advances in Neural Information Processing Systems}, volume~30, 2017.
\newblock URL \url{https://proceedings.neurips.cc/paper_files/paper/2017/hash/758a06618c69880a6cee5314ee42d52f-Abstract.html}.

\bibitem[Font et~al.(2025)Font, Alc{\'a}ntara-{\'A}vila, Rabault, Vinuesa, and Lehmkuhl]{font2025bubble}
Bernat Font, Francisco Alc{\'a}ntara-{\'A}vila, Jean Rabault, Ricardo Vinuesa, and Oriol Lehmkuhl.
\newblock Deep reinforcement learning for active flow control in a turbulent separation bubble.
\newblock \emph{Nature Communications}, 16\penalty0 (1):\penalty0 1422, 2025.
\newblock \doi{10.1038/s41467-025-56408-6}.

\bibitem[Fu et~al.(2023)Fu, Wu, Wang, Xie, Keten, G{\'o}mez-Bombarelli, and Jaakkola]{fu2023forces}
Xiang Fu, Zhenghao Wu, Wujie Wang, Tian Xie, Sinan Keten, Rafael G{\'o}mez-Bombarelli, and Tommi Jaakkola.
\newblock Forces are not enough: Benchmark and critical evaluation for machine learning force fields with molecular simulations.
\newblock \emph{Transactions on Machine Learning Research}, 2023.
\newblock URL \url{https://openreview.net/forum?id=A8pqQipwkt}.

\bibitem[Fujimoto \& Gu(2021)Fujimoto and Gu]{fujimoto2021minimalist}
Scott Fujimoto and Shixiang~Shane Gu.
\newblock A minimalist approach to offline reinforcement learning.
\newblock In \emph{Advances in Neural Information Processing Systems}, volume~34, 2021.
\newblock URL \url{https://papers.nips.cc/paper_files/paper/2021/hash/a8166da05c5a094f7dc03724b41886e5-Abstract.html}.

\bibitem[Gao et~al.(2024)Gao, Shi, Zhang, Wang, and Wu]{gao2024adaptive}
Feng Gao, Liangzhi Shi, Shenao Zhang, Zhaoran Wang, and Yi~Wu.
\newblock Adaptive-gradient policy optimization: Enhancing policy learning in non-smooth differentiable simulations.
\newblock In \emph{Proceedings of the 41st International Conference on Machine Learning}, volume 235 of \emph{Proceedings of Machine Learning Research}, pp.\  14844--14858. PMLR, 2024.
\newblock URL \url{https://proceedings.mlr.press/v235/gao24m.html}.

\bibitem[Guastoni et~al.(2023)Guastoni, Rabault, Schlatter, Azizpour, and Vinuesa]{guastoni2023deep}
Luca Guastoni, Jean Rabault, Philipp Schlatter, Hossein Azizpour, and Ricardo Vinuesa.
\newblock Deep reinforcement learning for turbulent drag reduction in channel flows.
\newblock \emph{The European Physical Journal E}, 46\penalty0 (4):\penalty0 27, 2023.
\newblock \doi{10.1140/epje/s10189-023-00285-8}.

\bibitem[Hafner et~al.(2020)Hafner, Lillicrap, Ba, and Norouzi]{hafner2020dream}
Danijar Hafner, Timothy Lillicrap, Jimmy Ba, and Mohammad Norouzi.
\newblock Dream to control: Learning behaviors by latent imagination.
\newblock In \emph{International Conference on Learning Representations}, 2020.
\newblock URL \url{https://openreview.net/forum?id=S1lOTC4tDS}.

\bibitem[Janner et~al.(2019)Janner, Fu, Zhang, and Levine]{janner2019when}
Michael Janner, Justin Fu, Marvin Zhang, and Sergey Levine.
\newblock When to trust your model: Model-based policy optimization.
\newblock In \emph{Advances in Neural Information Processing Systems (NeurIPS)}, volume~32, 2019.
\newblock URL \url{https://papers.nips.cc/paper_files/paper/2019/hash/5faf461eff3099671ad63c6f3f094f7f-Abstract.html}.

\bibitem[Lagemann et~al.(2025{\natexlab{a}})Lagemann, Mokbel, Gondrum, R{\"u}ttgers, Wang, Su{\'a}rez, Paehler, Bezgin, Buhendwa, Callaham, Ahnert, Zolman, Shao, Loiseau, Adams, Meinke, Schr{\"o}der, Lagemann, Lagemann, Vinuesa, and Brunton]{lagemann_2025_hydrogym_platform}
Christian Lagemann, Sajeda Mokbel, Miro Gondrum, Mario R{\"u}ttgers, Yuning Wang, Pol Su{\'a}rez, Ludger Paehler, Deniz~A. Bezgin, Aaron~B. Buhendwa, Jared~L. Callaham, Samuel Ahnert, Nicholas Zolman, Xiao Shao, Jean-Christophe Loiseau, Nikolaus Adams, Matthias Meinke, Wolfgang Schr{\"o}der, Kai Lagemann, Esther Lagemann, Ricardo Vinuesa, and Steven~L. Brunton.
\newblock The {HydroGym} reinforcement learning platform for fluid dynamics.
\newblock \emph{arXiv preprint arXiv:2512.17534}, 2025{\natexlab{a}}.
\newblock URL \url{https://arxiv.org/abs/2512.17534}.

\bibitem[Lagemann et~al.(2025{\natexlab{b}})Lagemann, Paehler, Callaham, Mokbel, Ahnert, Lagemann, Lagemann, Adams, and Brunton]{lagemann_2025_hydrogym}
Christian Lagemann, Ludger Paehler, Jared Callaham, Sajeda Mokbel, Samuel Ahnert, Kai Lagemann, Esther Lagemann, Nikolaus Adams, and Steven Brunton.
\newblock {HydroGym}: A reinforcement learning platform for fluid dynamics.
\newblock In \emph{Proceedings of the 7th Annual Learning for Dynamics \& Control Conference}, volume 283 of \emph{Proceedings of Machine Learning Research}, pp.\  497--512. PMLR, 2025{\natexlab{b}}.
\newblock URL \url{https://proceedings.mlr.press/v283/lagemann25a.html}.

\bibitem[Lambert et~al.(2020)Lambert, Amos, Yadan, and Calandra]{lambert2019objective}
Nathan Lambert, Brandon Amos, Omry Yadan, and Roberto Calandra.
\newblock Objective mismatch in model-based reinforcement learning.
\newblock In \emph{Proceedings of the 2nd Conference on Learning for Dynamics and Control}, volume 120 of \emph{Proceedings of Machine Learning Research}, pp.\  761--770. PMLR, 2020.
\newblock URL \url{https://proceedings.mlr.press/v120/lambert20a.html}.

\bibitem[Liu \& MacArt(2024)Liu and MacArt]{liu2024adjoint}
Xuemin Liu and Jonathan~F. MacArt.
\newblock Adjoint-based machine learning for active flow control.
\newblock \emph{Physical Review Fluids}, 9\penalty0 (1):\penalty0 013901, 2024.
\newblock \doi{10.1103/PhysRevFluids.9.013901}.

\bibitem[Metz et~al.(2021)Metz, Freeman, Schoenholz, and Kachman]{metz2021gradients}
Luke Metz, C.~Daniel Freeman, Samuel~S. Schoenholz, and Tal Kachman.
\newblock Gradients are not all you need.
\newblock \emph{arXiv preprint arXiv:2111.05803}, 2021.
\newblock URL \url{https://arxiv.org/abs/2111.05803}.

\bibitem[Nesterov(2018)]{nesterov2018lectures}
Yurii Nesterov.
\newblock \emph{Lectures on Convex Optimization}, volume 137 of \emph{Springer Optimization and Its Applications}.
\newblock Springer, Cham, second edition, 2018.
\newblock \doi{10.1007/978-3-319-91578-4}.

\bibitem[Paris et~al.(2021)Paris, Beneddine, and Dandois]{paris2021robust}
Romain Paris, Samir Beneddine, and Julien Dandois.
\newblock Robust flow control and optimal sensor placement using deep reinforcement learning.
\newblock \emph{Journal of Fluid Mechanics}, 913:\penalty0 A25, 2021.
\newblock \doi{10.1017/jfm.2020.1170}.

\bibitem[Parmas et~al.(2018)Parmas, Rasmussen, Peters, and Doya]{parmas2018pipps}
Paavo Parmas, Carl~Edward Rasmussen, Jan Peters, and Kenji Doya.
\newblock {PIPPS}: Flexible model-based policy search robust to the curse of chaos.
\newblock In \emph{Proceedings of the 35th International Conference on Machine Learning}, volume~80 of \emph{Proceedings of Machine Learning Research}, pp.\  4065--4074. PMLR, 2018.
\newblock URL \url{https://proceedings.mlr.press/v80/parmas18a.html}.

\bibitem[Psenka et~al.(2024)Psenka, Escontrela, Abbeel, and Ma]{psenka2024qsm}
Michael Psenka, Alejandro Escontrela, Pieter Abbeel, and Yi~Ma.
\newblock Learning a diffusion model policy from rewards via {Q}-score matching.
\newblock In \emph{Proceedings of the 41st International Conference on Machine Learning}, volume 235 of \emph{Proceedings of Machine Learning Research}, pp.\  41163--41182. PMLR, 2024.
\newblock URL \url{https://proceedings.mlr.press/v235/psenka24a.html}.

\bibitem[Rabault et~al.(2019)Rabault, Kuchta, Jensen, R{\'e}glade, and Cerardi]{rabault2019artificial}
Jean Rabault, Miroslav Kuchta, Atle Jensen, Ulysse R{\'e}glade, and Nicolas Cerardi.
\newblock Artificial neural networks trained through deep reinforcement learning discover control strategies for active flow control.
\newblock \emph{Journal of Fluid Mechanics}, 865:\penalty0 281--302, 2019.
\newblock \doi{10.1017/jfm.2019.62}.

\bibitem[Sato \& Schmidt(2025)Sato and Schmidt]{sato_2025_parametric_pod}
Shintaro Sato and Oliver~T. Schmidt.
\newblock Parametric reduced-order modelling and mode sensitivity of actuated cylinder flow from a matrix manifold perspective.
\newblock \emph{Journal of Fluid Mechanics}, 1021:\penalty0 A44, 2025.
\newblock \doi{10.1017/jfm.2025.10733}.

\bibitem[Suh et~al.(2022)Suh, Simchowitz, Zhang, and Tedrake]{suh2022differentiable}
Hyung~Ju Suh, Max Simchowitz, Kaiqing Zhang, and Russ Tedrake.
\newblock Do differentiable simulators give better policy gradients?
\newblock In \emph{Proceedings of the 39th International Conference on Machine Learning}, volume 162 of \emph{Proceedings of Machine Learning Research}, pp.\  20668--20696. PMLR, 2022.
\newblock URL \url{https://proceedings.mlr.press/v162/suh22b.html}.

\bibitem[Vernade et~al.(2026)Vernade, Eberhard, White, D{\"o}rfler, Szepesv{\'a}ri, Krstic, and Muehlebach]{vernade2026foundations}
Claire Vernade, Onno Eberhard, Martha White, Florian D{\"o}rfler, Csaba Szepesv{\'a}ri, Miroslav Krstic, and Michael Muehlebach.
\newblock Foundations of reinforcement learning and control: Connections and new perspectives.
\newblock \emph{arXiv preprint arXiv:2608.02433}, 2026.
\newblock URL \url{https://arxiv.org/abs/2608.02433}.

\bibitem[Vinuesa \& Brunton(2022)Vinuesa and Brunton]{vinuesa2022deep}
Ricardo Vinuesa and Steven~L Brunton.
\newblock Enhancing computational fluid dynamics with machine learning.
\newblock \emph{Nature Computational Science}, 2:\penalty0 358--366, 2022.
\newblock \doi{10.1038/s43588-022-00264-7}.

\bibitem[Voelcker et~al.(2022)Voelcker, Liao, Garg, and Farahmand]{voelcker2023vagram}
Claas~A. Voelcker, Victor Liao, Animesh Garg, and Amir-massoud Farahmand.
\newblock Value gradient weighted model-based reinforcement learning.
\newblock In \emph{International Conference on Learning Representations}, 2022.
\newblock URL \url{https://openreview.net/forum?id=4-D6CZkRXxI}.

\bibitem[Wang et~al.(2019)Wang, Bao, Clavera, Hoang, Wen, Langlois, Zhang, Zhang, Abbeel, and Ba]{langlois2019benchmarking}
Tingwu Wang, Xuchan Bao, Ignasi Clavera, Jerrick Hoang, Yeming Wen, Eric Langlois, Shunshi Zhang, Guodong Zhang, Pieter Abbeel, and Jimmy Ba.
\newblock Benchmarking model-based reinforcement learning.
\newblock \emph{arXiv preprint arXiv:1907.02057}, 2019.
\newblock URL \url{https://arxiv.org/abs/1907.02057}.

\bibitem[Weiner \& Geise(2024)Weiner and Geise]{weiner_2024_model_based_drl}
Andre Weiner and Janis Geise.
\newblock Model-based deep reinforcement learning for accelerated learning from flow simulations.
\newblock \emph{arXiv preprint arXiv:2402.16543}, 2024.
\newblock URL \url{https://arxiv.org/abs/2402.16543}.

\bibitem[Yang et~al.(2023)Yang, Huang, Lei, Zhong, Yang, Fang, Wen, Zhou, and Lin]{yang2023dipo}
Long Yang, Zhixiong Huang, Fenghao Lei, Yucun Zhong, Yiming Yang, Cong Fang, Shiting Wen, Binbin Zhou, and Zhouchen Lin.
\newblock Policy representation via diffusion probability model for reinforcement learning.
\newblock \emph{arXiv preprint arXiv:2305.13122}, 2023.
\newblock URL \url{https://arxiv.org/abs/2305.13122}.

\bibitem[Yu et~al.(2020)Yu, Thomas, Yu, Ermon, Zou, Levine, Finn, and Ma]{yu2020mopo}
Tianhe Yu, Garrett Thomas, Lantao Yu, Stefano Ermon, James~Y. Zou, Sergey Levine, Chelsea Finn, and Tengyu Ma.
\newblock {MOPO}: Model-based offline policy optimization.
\newblock In \emph{Advances in Neural Information Processing Systems}, volume~33, 2020.
\newblock URL \url{https://proceedings.neurips.cc/paper/2020/hash/a322852ce0df73e204b7e67cbbef0d0a-Abstract.html}.

\bibitem[Zamora~Mora et~al.(2021)Zamora~Mora, Peychev, Ha, Vechev, and Coros]{mora2021pods}
Miguel~Angel Zamora~Mora, Momchil Peychev, Sehoon Ha, Martin Vechev, and Stelian Coros.
\newblock {PODS}: Policy optimization via differentiable simulation.
\newblock In \emph{Proceedings of the 38th International Conference on Machine Learning}, volume 139 of \emph{Proceedings of Machine Learning Research}, pp.\  7805--7817. PMLR, 2021.
\newblock URL \url{https://proceedings.mlr.press/v139/mora21a.html}.

\bibitem[Zhang et~al.(2024)Zhang, Yousif, Xu, Zhou, Yu, and Lim]{zhang_2024_surrogate_drl_square_cylinder}
Meng Zhang, Mustafa~Z. Yousif, Minze Xu, Haifeng Zhou, Linqi Yu, and HeeChang Lim.
\newblock Efficient active flow control strategy for confined square cylinder wake using deep learning-based surrogate model and reinforcement learning.
\newblock \emph{arXiv preprint arXiv:2408.14232}, 2024.
\newblock URL \url{https://arxiv.org/abs/2408.14232}.

\bibitem[Zhou et~al.(2025)Zhou, Zhang, and Zhu]{zhou2025channel}
Zisong Zhou, Mengqi Zhang, and Xiaojue Zhu.
\newblock Reinforcement-learning-based control of turbulent channel flows at high {R}eynolds numbers.
\newblock \emph{Journal of Fluid Mechanics}, 1006:\penalty0 A12, 2025.
\newblock \doi{10.1017/jfm.2025.27}.

\bibitem[Zolman et~al.(2025)Zolman, Lagemann, Fasel, Kutz, and Brunton]{zolman_2024_sindy_rl}
Nicholas Zolman, Christian Lagemann, Urban Fasel, J.~Nathan Kutz, and Steven~L. Brunton.
\newblock {SINDy-RL} for interpretable and efficient model-based reinforcement learning.
\newblock \emph{Nature Communications}, 16:\penalty0 10714, 2025.
\newblock \doi{10.1038/s41467-025-65738-4}.

\end{thebibliography}

\appendix
\renewcommand{\topfraction}{0.95}
\renewcommand{\bottomfraction}{0.95}
\renewcommand{\textfraction}{0.05}
\renewcommand{\floatpagefraction}{0.80}
\makeatletter
\setlength{\@fptop}{0pt}
\setlength{\@fpsep}{14pt plus 0fil}
\setlength{\@fpbot}{0pt plus 1fil}
\makeatother
\section{Scope and evidence rules}
\label{app:scope}

The protocols, null results and uncertainty units behind the main paper are
recorded below, under four rules.

\begin{enumerate}
    \item A prediction score measures forward model accuracy.
    \item A gradient audit measures local action sensitivity on excluded states.
    \item Controller confirmation uses initial conditions excluded from selection.
    \item Only results of completed runs enter the paper, and comparisons without a completed run are not reported.
\end{enumerate}

A rule is predefined when it was fixed and recorded before the measurement it governs. Every
threshold, selection rule, success rule, and stopping rule quoted below is predefined in that
sense, and any quantity that is not is called exploratory where it appears.

Controller performance is always measured in the high-fidelity solver. The
independent unit is stated for every result. It can be a training seed, a
paired initial condition, or both. Each rollout is one trajectory from one initial condition.

The number of independent units differs across the test cases and is stated for each row. On the rotating cylinder, the released
controller uses four training seeds and eight paired initial conditions per seed,
and all four seed means are positive. The extracted harmonic law is deterministic conditional
on a frozen audited surrogate, while three independently fitted pipeline
surrogates extract phases of $143.5^\circ$, $164.3^\circ$, and $146.0^\circ$. By comparison, a harmonic fitted directly to the eight measured gradients varies within $4.2^\circ$ when any one state is left out. On the three cylinder cluster, the anchored and released pathwise controllers each use four training seeds with eight initial conditions per seed. The matched
model-free baseline uses three training seeds on the same eight conditions. Forced two-dimensional turbulence uses eight controller training seeds, fixed before training, and evaluates each on eight held-out initial conditions. Each final chaotic-equation comparison uses
eight held-out fields at each forcing amplitude.

The model-free baselines are proximal policy optimization, or PPO, on the two cylinder
cases and on forced turbulence, and twin delayed deep deterministic policy gradient, or TD3, on the
chaotic equation. They differ in provenance and are not interchangeable. The cylinder,
three cylinder, and forced-turbulence baselines are trained and evaluated in the solvers
used here. The chaotic-equation comparisons use matched local TD3 and polynomial TD3
checkpoints evaluated on the same held-out fields as the surrogate-based controllers. Therefore, each row supports a comparison only within its own test case and protocol.

\section{Acceptance criteria and threshold sensitivity}
\label{app:thresholds}

No recomputed acceptance decision changes over a wide interval around any of the three thresholds.
The predefined acceptance thresholds are $\tau_1=0.70$ for pooled correlation,
$\tau_2=0.70$ for sign agreement, and $\tau_3=0.50$ for median per-actuator correlation, in the
notation of the main paper. The recomputation uses the stored audit statistics and no new solver evaluation. It covers three groups of decisions. The rejected and the accepted three cylinder surrogates and the five chaotic-equation audits use all three statistics. The rotating cylinder audit at Reynolds number 200 uses the pooled correlation and the sign agreement only, because the cylinder has one actuator. The controller check of the eight forced-turbulence controllers uses the two conditions defined in Appendix~\ref{app:turbulence}. The rotating cylinder audit at Reynolds number 100, the forced turbulence audit and the three deployed-action audits at $\mu=0.121$ are not in the table, and they also keep their decisions over the tested range. Table~\ref{tab:threshold-sensitivity} gives the largest one-axis
interval within the tested grid, $0.50$ to $0.90$ for the first two thresholds and $0.30$ to $0.80$ for the third, over which no reported decision changes when the other two thresholds remain at
their stated values. Each of these decisions also remains unchanged when all three thresholds
vary together over the grid-aligned box formed by these intervals.

\begin{table}[htbp]
\centering
\caption{No reported audit decision changes over a wide interval around each threshold. The intervals measure the stability of each decision and do not derive the thresholds.}
\label{tab:threshold-sensitivity}
\begin{tabular}{lcc}
\toprule
Audit statistic & Threshold & Interval preserving every decision \\
\midrule
Pooled correlation & $\tau_1=0.70$ & $[0.50,0.90]$ \\
Sign agreement & $\tau_2=0.70$ & $[0.50,0.89]$ \\
Median per-actuator correlation & $\tau_3=0.50$ & $[0.30,0.63]$ \\
\bottomrule
\end{tabular}
\end{table}

The three-dimensional stability box runs from $(0.50,0.50,0.30)$ to
$(0.90,0.89,0.63)$. Therefore, the reported decisions do not depend on a single narrow cutoff. Because the analysis is retrospective and uses a finite grid, it
does not claim that every value inside the continuous box is scientifically equivalent.
It reports only that the stored decisions do not change over the tested range.

The thresholds are empirical, and no theoretical result links them to the performance of a trained controller. Neither of two alternative statistics would make the audit more reliable than the median per-actuator correlation, and Table~\ref{tab:alt-statistics} lists both. The median is used in place of the minimum because, with few held-out states, a single noisy actuator correlation would otherwise decide the audit alone. Replacing the median per-actuator correlation by the minimum over actuators, at the same threshold of $0.50$, leaves the four decisions in the table unchanged. The rejected force model still fails on its pooled correlation and sign agreement, the accepted latent field model and the chaotic-equation gradient model still pass, and the forced turbulence surrogate fails under either statistic. The minimum would, however, reject two audits not listed there, the second audits of the two chaotic-equation gradient models, on states reached by controllers built on them, where one of the eight actuators has a negative correlation, $-0.78$ at $\mu=0.225$ and $-0.89$ at $\mu=0.121$. The median per-state cosine $\cos\theta$ defined in Section 3.2 of the main paper, compared with $0.70$, would change one decision, because the forced turbulence surrogate would pass. The pooled correlation and the median per-actuator correlation are kept, because the per-state cosine is dominated by the part of the gradient shared by all states and consequently can accept a surrogate whose state-to-state variation is wrong, as it is on forced turbulence.

\begin{table}[htbp]
\centering
\caption{Replacing the median per-actuator correlation by the minimum changes none of the four decisions listed, while a threshold on the median per-state cosine would accept the forced turbulence surrogate. Values are computed from the stored audit gradients. }
\label{tab:alt-statistics}
\small
\begin{tabularx}{\textwidth}{>{\raggedright\arraybackslash}X>{\centering\arraybackslash}p{0.15\textwidth}>{\centering\arraybackslash}p{0.15\textwidth}>{\centering\arraybackslash}p{0.15\textwidth}}
\toprule
Surrogate & Median actuator correlation & Minimum actuator correlation & Median per-state cosine \\
\midrule
three cylinder force model, rejected & $0.512$ & $0.370$ & $0.548$ \\
three cylinder latent field model, new held-out states & $0.638$ & $0.548$ & $1.000$ \\
forced turbulence neural operator & $-0.342$ & $-0.648$ & $0.859$ \\
chaotic equation gradient model, $\mu=0.225$ & $0.890$ & $0.804$ & $0.990$ \\
\bottomrule
\end{tabularx}
\end{table}

The audit decisions do not depend on the step $\epsilon$ of the central finite differences. The step was varied on the two test cases with inexpensive solvers. On the parametric chaotic equation, the eight audited states were recomputed with the same policy for the $H-1$ actions that follow the perturbed first action, the same horizon and the same seeds at $\epsilon=0.025$, $0.05$ and $0.10$. On forced turbulence the 20 audited states were recomputed at the same three values. The recomputation at the step used in the paper, $\epsilon=0.05$, reproduces the stored gradients to a relative error below $10^{-5}$. Halving or doubling the step changes the gradient vectors by a median relative difference of at most $0.004\%$ on the chaotic equation and $0.04\%$ on forced turbulence, with sign agreement $1.00$ in both cases. Every audit statistic changes by less than $10^{-3}$, so the chaotic-equation gradient model still passes at pooled correlation $0.979$ and the forced turbulence surrogate still fails at $0.530$. The rotating cylinder and the three cylinder cluster were not rechecked, because each recomputation costs several hours of solver time.

\section{State dependence of the measured gradient}
\label{app:feedback-content}

The finite-difference campaigns also measure how far each gradient field depends on the state, through the state-independent fraction $f$ of Equation~(7) of the main paper. Table~\ref{tab:feedback-content} collects those values for every test case. Each entry describes the field over the states that were sampled, around the reference action at which they were sampled, so one test case can have several values.

\begin{table}[htbp]
\centering
\caption{The state dependence of the measured gradient field runs from total to almost none across
these conditions. Each value of $f$ is computed from the high-fidelity finite differences at the
listed action coordinates and reference action. Each interval is a percentile stability interval
from $10{,}000$ resamples of the sampled states, or of the sources for the three deployed-action rows. It shows how far the value moves under resampling and is not a confidence interval. The rotating cylinder entry uses eight shedding-phase bins fixed by construction, so no resampling interval applies to it.}
\label{tab:feedback-content}
\begin{tabularx}{\textwidth}{>{\raggedright\arraybackslash}X rrr}
\toprule
flow and sampled action coordinates & states & $f$ & stability interval \\
\midrule
rotating cylinder, one rotation & 8 & $0.000$ & fixed phase bins \\
three cylinder cluster, three rotations & 16 & $0.838$ & $[0.808,\,0.903]$ \\
forced turbulence, four forcing modes & 20 & $0.654$ & $[0.561,\,0.831]$ \\
parametric chaotic equation, $\mu=0.225$, fixed reference action & 24 & $0.986$ & $[0.978,\,0.996]$ \\
parametric chaotic equation, $\mu=0.121$, actions of the first controller built at $\mu=0.121$ & 24 & $0.028$ & $[0.017,\,0.164]$ \\
parametric chaotic equation, $\mu=0.121$, actions of the second controller built at $\mu=0.121$ & 24 & $0.029$ & $[0.029,\,0.154]$ \\
parametric chaotic equation, $\mu=0.121$, actions of the controller transferred from $\mu=0.225$ & 24 & $0.024$ & $[0.020,\,0.094]$ \\
\bottomrule
\end{tabularx}
\end{table}

\section{Proofs}
\label{app:proofs}

The derivations that the main paper states without proof are collected here.

\subsection{Derivation of the local descent bound}

A step against the surrogate gradient lowers the high-fidelity objective by at least the amount the
following inequality certifies.

\begin{proof}
Let $\mathbf s=-\eta\widehat{\mathbf g}$. The fundamental theorem of calculus gives
\begin{equation*}
J_H(\mathbf x,\mathbf a+\mathbf s)-J_H(\mathbf x,\mathbf a)
=\int_0^1 \nabla_{\mathbf a} J_H(\mathbf x,\mathbf a+t\mathbf s)^\top \mathbf s\,dt.
\end{equation*}
Add and subtract $\mathbf g$ inside the integral. The Lipschitz assumption and the
Cauchy Schwarz inequality give
\begin{equation*}
J_H(\mathbf x,\mathbf a+\mathbf s)-J_H(\mathbf x,\mathbf a)
\leq \mathbf g^\top \mathbf s+\int_0^1 Lt\lVert \mathbf s\rVert^2\,dt
=-\eta \mathbf g^\top\widehat{\mathbf g}
+\frac{L\eta^2}{2}\lVert\widehat{\mathbf g}\rVert^2.
\end{equation*}
The step size condition of Equation~(9) of the main paper makes the final
quantity negative.
\end{proof}

\subsection{Scope of the directional bound}

Three limits restrict how the descent bound may be used. It holds at one state while pooled correlation
aggregates over states and actuators, so a pooled value above the threshold leaves the condition
unestablished at any particular state. Sign agreement alone does not bound the cosine. One dominant coordinate whose sign disagrees can pull the cosine below zero while most signs still match. The result also bounds a single update at one state. Extending it along a
trajectory would need a measured smoothness constant, and the anchor penalty replaces that constant
with an empirical restriction on how far the policy may move. Therefore, closed-loop performance, controller selection and anchor release are tested in the high-fidelity solver.

\subsection{Proof of Proposition 1}

The ratio of the two certified decreases equals the squared cosine of the angle between the gradients.

\begin{proof}
The bound derived above is
$-\eta\,\mathbf g^\top\widehat{\mathbf g}+\tfrac{L\eta^2}{2}\lVert\widehat{\mathbf g}\rVert^2$. It is a
convex quadratic in $\eta$ and is minimized at
$\eta^\star=\mathbf g^\top\widehat{\mathbf g}/(L\lVert\widehat{\mathbf g}\rVert^2)$. This step size is positive,
because $\mathbf g^\top\widehat{\mathbf g}>0$ by assumption, and it lies in the admissible step size interval of
Equation~(9) of the main paper. Substituting
$\eta^\star$ gives
$-(\mathbf g^\top\widehat{\mathbf g})^2/(2L\lVert\widehat{\mathbf g}\rVert^2)
=-\lVert \mathbf g\rVert^2\cos^2\theta/(2L)$.
Setting $\widehat{\mathbf g}=\mathbf g$ gives $-\lVert \mathbf g\rVert^2/(2L)$. The ratio is
$\cos^2\theta$ and is independent of $L$ and of $\lVert\widehat{\mathbf g}\rVert$.
The minimizing step is an oracle quantity because it depends on the true inner product and on
$L$. The ratio compares two local quadratic bounds. The ratio is not realized controller improvement.
The reported experiments use pooled Pearson correlation, sign agreement, and median per-actuator Pearson correlation. Those aggregate statistics are not substituted for the uncentered
per-state cosine in this proof.
\end{proof}

\section{Surrogate implementation}
\label{app:surrogate}

\begin{figure}[htbp]
\centering
\includegraphics[width=0.98\textwidth]{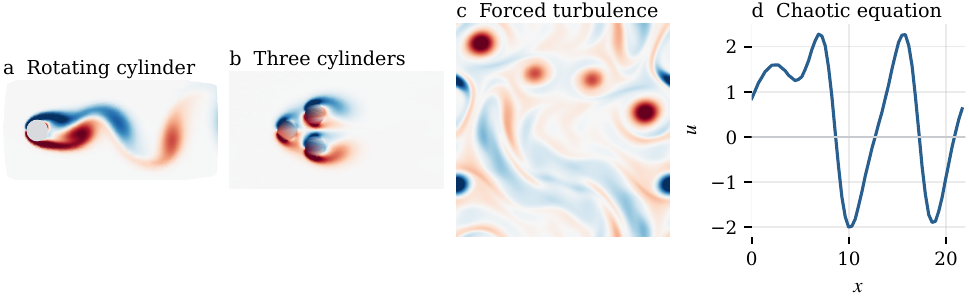}
\caption{The four control problems differ in dimension, in boundary condition, and in the
structure the controller must act on. The panels show the
rotating cylinder wake, the three cylinder cluster, forced two-dimensional
turbulence, and the one-dimensional chaotic equation. Panels a to c show vorticity from the high-fidelity solver, red positive and blue negative, and panel d shows the field $u$ along $x$.}
\label{fig:supp-flow-context}
\end{figure}

Normalization statistics are fitted on training states and then frozen. Every
data split keeps all samples from one high-fidelity state together. This prevents
paired action perturbations from the same state from entering both fitting and
evaluation sets. The exception is the first audit of the three cylinder latent field model, which Appendix~\ref{app:pinball} describes. Rollout models use multi-step training. Their predicted
observations and objective quantities are compared with high-fidelity solver data
at every step. Gradient supervision is used on the surrogates whose fitting data include paired high-fidelity
labels, as stated for each case below. Figure~\ref{fig:supp-flow-context} shows a representative field of each test case.

\subsection{Rotating cylinder wake}

The selected model reads vorticity on 8,744 mesh nodes. Each node has 12 input
features. These features contain normalized vorticity, position, requested
action, filtered actuator state, measured lift, drag, wall shear, a wall mask,
and signed distance information. A Transolver encoder has hidden width 96,
eight attention heads, 24 learned slices, and four layers. It maps the field
to a 128-component latent state. A multilayer perceptron with hidden width 256
advances this state using the action. Lift, drag, wall shear, and reward are
decoded over a 16-step rollout.

The model uses AdamW with learning rate $3\times10^{-4}$, weight decay
$10^{-5}$, batches of 16, and gradient norm limit 1. Training runs for at most
30 epochs with early stopping after 15 epochs without improvement. The final
action-Jacobian correction starts from the fixed 16-step model and uses paired
high-fidelity action responses at horizon 25. The correction compares normalized
gradient direction and action ranking. Its weight is 0.5. It is the cylinder form of the gradient term of the model loss in the main text.

\begin{figure}[htbp]
\centering
\includegraphics[width=\textwidth]{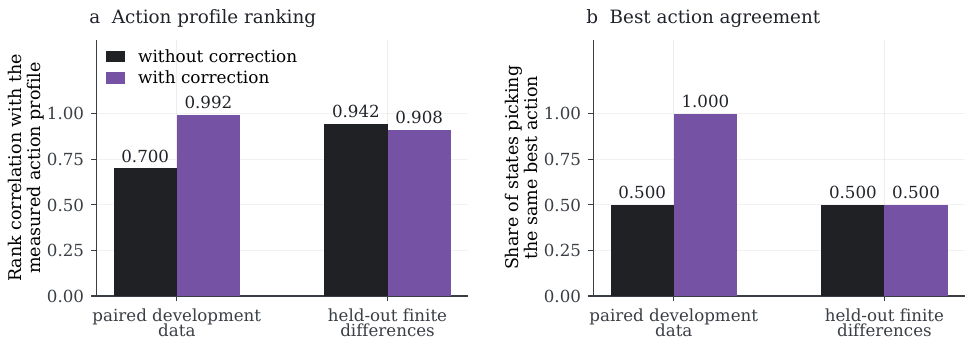}
\caption{The action-Jacobian correction improves the development metrics and does not improve the
held-out ones.
The correction improves the action profile metrics on the paired development
data. It does not improve rank correlation or best action agreement on the
held-out finite differences. Both models use one matched training seed.}
\label{fig:supp-cylinder-correction-ablation}
\end{figure}

A matched one-seed ablation uses the same architecture, base data and optimizer with and without this correction, and both models are trained from scratch for 60 epochs. Figure~\ref{fig:supp-cylinder-correction-ablation} shows the result. On the paired
development data, objective profile rank correlation changes from 0.700 to
0.992 and argmin agreement changes from 0.50 to 1.00. On the independent
held-out finite differences, rank correlation changes from 0.942 to 0.908 and argmin
agreement remains 0.50. The correction has no established generalization
benefit in this ablation.

\subsection{Three cylinder cluster}

Each field snapshot contains velocity, pressure, and vorticity on 25,695 mesh
nodes. The node features and coordinates are standardized with statistics from
the fitting rollouts. The encoder has the same hidden width 96, eight heads,
24 slices, and four layers. Its 128-component latent state is advanced by a
multilayer perceptron with hidden width 256. Decoders predict the nine measured
lift, drag, and actuator quantities and the objective over 16 steps.

The 12 high-fidelity rollouts are split by complete rollout. They produce 648
training windows and 216 validation windows. AdamW uses learning rate
$3\times10^{-4}$, weight decay $10^{-5}$, batches of 2, and gradient norm limit
1. Training runs for at most 80 epochs with patience 18. Paired gradient labels
are divided into 72 fitting and 72 held-out gradient labels, each the three-component gradient at one state and one reference action. One component of one label is one comparison, so 72 labels give 216 comparisons. A second audit uses 48 comparisons on 16 additional held-out states.

\subsection{Forced two-dimensional turbulence}

The high-fidelity solver is pseudo-spectral: it computes derivatives in Fourier space and the nonlinear term in physical space. In that product, the highest third of the resolved wavenumbers is set to zero, the two-thirds rule, which removes the aliasing error that the product would otherwise fold back onto the resolved scales. Time is advanced with a Runge--Kutta scheme for the nonlinear term and a Crank--Nicolson scheme for the viscous term, with time step $0.001$ and 100 solver steps per control step, on a $256$ by $256$ grid at $Re=200$.

The field models read normalized periodic vorticity and four forcing
amplitudes. The reported surrogate is the full-resolution Fourier neural operator, which predicts vorticity directly and fails the audit, with pooled correlation $0.53$, sign agreement $0.55$ and median per-actuator correlation $-0.34$. A latent version, described next, was also trained. The latent version uses four Fourier layers with width 32 and 16
spectral modes in each direction. Global mean, standard deviation, maximum,
and minimum features are mapped to a 192-component latent state. A two layer
network with hidden width 256 advances that state. Separate heads predict
enstrophy and the control objective.

The latent model uses 16-step training and a 20-step gradient audit. At most
1,800 windows are used. $15\%$ are reserved for validation. AdamW uses
learning rate $3\times10^{-4}$, batches of 16, weight decay $10^{-4}$, and at
most 40 epochs. The full-resolution Fourier neural operator was trained before this study by repeatedly deploying a controller and refitting the surrogate, and the data of those iterations were not kept.

The neural policy for forced turbulence takes 64 complex Fourier coefficients of the vorticity field at the lowest wavenumbers, eight along each axis, starting from zero along one axis and centered on zero along the other. These coefficients give 128 real inputs, and the network has two hidden layers of 128 units with GELU activations. It outputs the three action components orthogonal to the extracted law, scaled so that the action stays within five degrees of the direction of the law.

\subsection{Parametric chaotic equation}

The benchmark solves Equation~(3) of the main paper on a periodic domain of length 22 with 64 grid points and a time step of $0.1$. Each control step spans two time steps, control starts at time 100, and an episode lasts 999 control steps. Eight actuators apply Gaussian forcing profiles centered every eighth grid point, and eight point sensors sit at the actuator centers. The transition model described here is the transition surrogate of the first stage in Section 4.4 of the main paper. The selected model receives eight sensor values, the equation parameter
$\mu$, and eight actuator values. An encoder with hidden width 128 maps the
nine-component observation to a 32-component latent state. Two hidden layers
advance the latent state. A decoder predicts the next observation. A separate
positive output head predicts the state cost.

The initial data contain 35,964 transitions. $90\%$ are used for
fitting and $10\%$ for validation. AdamW uses learning rate
$5\times10^{-4}$, weight decay $10^{-5}$, batches of 512, and at most 80
epochs. The model is then fine-tuned on 50,964 transitions, which add 7,500 transitions collected in closed loop and counted twice. The fine-tuning includes a loss on paired high-fidelity gradient labels at horizon 5, with weight $0.25$, on 92 states, and the audit labels are excluded from this loss. The fine-tuned model passes its own audit at horizon 5, with pooled correlation $0.982$, sign agreement $0.934$ and median per-actuator correlation $0.987$ on 80 held-out states. Controller
evaluation also uses states reached during high-fidelity deployment to test the
change in state distribution.

\section{Rotating cylinder wake evidence}
\label{app:cylinder}

The rotating cylinder supplies the paper's constructive case. It also supplies the clearest negative result, because optimization through an accurate surrogate, without the anchor penalty and from an initialization not derived from the law, makes the objective worse.

\subsection{Evaluation protocol}

One objective, one action bound, and one solver setting serve every rotating cylinder result in this paper. The cylinder objective is
\[
J=\overline{C_D}+0.2\,\overline{|C_L|}.
\]
The lift term prevents steady rotation from appearing useful only because of
drag reduction. All evaluations use the medium-mesh HydroGym direct numerical simulation. Each Reynolds number uses its own uncontrolled baseline. The action bound is $|a|\leq\pi/2$. Table~\ref{tab:supp-cylinder} lists every rotating cylinder controller, and Appendix~\ref{sup:noise} tests the extracted law under sensor noise.

\begin{table}[htbp]
\centering
\caption{The extracted law and the released pathwise controller at $Re=100$, and the native extracted law and native anchored pathwise controller at $Re=200$, are the rotating cylinder rows fixed before evaluation. Uncertainty is a sample standard deviation over the unit stated in the last column. Exploratory rows are
not reported as confirmed controller results. A row marked reference is a baseline that the other rows are compared with. The fixed checkpoint is the PPO controller at the end of its interaction budget, and the peak checkpoint is the best scoring
checkpoint of each seed, which uses the evaluation history and is consequently selection biased.}
\begin{tabularx}{\textwidth}{lrrX}
\toprule
Result & Re & Gain (\%) & Statistical unit and status \\
\midrule
Phase polynomial pathwise & 100 & $1.69\pm0.21$ & 24 paired rollouts, three training seeds, best high-fidelity iteration, exploratory \\
Model-free PPO, fixed checkpoint & 100 & $0.21\pm3.54$ & 12 paired rollouts, three seeds $\times$ four initial conditions, reference \\
Model-free PPO, peak checkpoint & 100 & $2.24\pm0.80$ & three seeds, best of three checkpoints each, selection biased reference \\
Extracted law, $164^\circ$ & 100 & $3.83\pm0.14$ & eight paired initial conditions, confirmed extraction \\
Phase line search, $200^\circ$ & 100 & $4.95\pm0.13$ & eight paired initial conditions, exploratory \\
Released pathwise & 100 & $3.33\pm0.15$ & four seeds $\times$ eight initial conditions \\
Latent PPO & 200 & $6.78\pm3.21$ & model-based, three training seeds, best high-fidelity iteration scored against the mean uncontrolled objective, selection biased \\
Model-free PPO, fixed checkpoint & 200 & $3.93\pm7.03$ & three seeds, reference \\
Model-free PPO, peak checkpoint & 200 & $7.63\pm1.43$ & three seeds, best checkpoint each, selection biased reference \\
Native extracted law, $151.7^\circ$ & 200 & $3.46\pm0.21$ & eight paired initial conditions, native surrogate extraction \\
Native anchored pathwise & 200 & $2.98\pm0.25$ & eight initial conditions, one training seed, confirmed high-fidelity evaluation \\
Amplitude-selected law & 200 & $13.66\pm0.06$ & eight initial conditions, exploratory amplitude selection \\
Exploratory-anchor pathwise & 200 & $13.40\pm0.20$ & four seeds $\times$ eight initial conditions, anchor inherited from the exploratory high-fidelity selection \\
\bottomrule
\end{tabularx}
\label{tab:supp-cylinder}
\end{table}

The released controller is not described as unguided discovery. It was
initialized by law projection and was trained with an
anchor before that anchor was reduced to zero.

The main text reports the rotating cylinder at $Re=100$ only, together with the frozen transfer
of that test case's law to higher Reynolds numbers. A separate campaign trained new surrogates natively at $Re=200$. Its full record appears below and
none of it appears in the main text, because it adds a second controller development sequence to a
flow that already supplies the paper's constructive case.

The controller rows use different anchor targets and consequently support different conclusions.
At $Re=100$ the anchor target is the audited $164^\circ$ law
at amplitude 0.2, fixed by the gradient audit before high-fidelity evaluation. At
$Re=200$ the native pipeline extracts $151.7^\circ$ at amplitude 0.2 from three freshly
trained surrogates. Its extracted law gives $3.46\pm0.21\%$, and its pathwise controller, trained with one seed, gives $2.98\pm0.25\%$, each a sample standard deviation over eight paired initial conditions. These are the native Reynolds number 200 results.

The three native surrogates predict the measured action gradient accurately. An audit at zero action, on eight states spread over one shedding cycle of the uncontrolled flow and over 25 control steps, gives the following values. The ensemble prediction reaches
Pearson correlation $0.942$, sign agreement $1.00$, root mean square error $0.0448$, and a
measured-on-predicted through-origin slope of $1.03$. The individual surrogate correlations
are $0.931$, $0.968$, and $0.925$. Every high-fidelity state matches its frozen reference field
exactly by hash. Since the local gradient estimate is accurate, the weak native controller is attributed mainly to the extraction step or to sustained deployment.

The stronger Reynolds number 200 controller uses the exploratory $230^\circ$ law at amplitude 0.4,
selected by the high-fidelity phase and amplitude search. Therefore, the pathwise row inherits a high-fidelity selection and slightly
undershoots the selected law, $13.40\pm0.20\%$ over 32 paired rollouts against $13.66\pm0.06\%$ over eight initial conditions, each a sample standard deviation. It
demonstrates that anchored release retains a law selected in the solver across four
training seeds. The audit did not discover that law independently.

\subsection{Mechanism and falsification controls}

\begin{figure}[htbp]
\centering
\includegraphics[width=\textwidth]{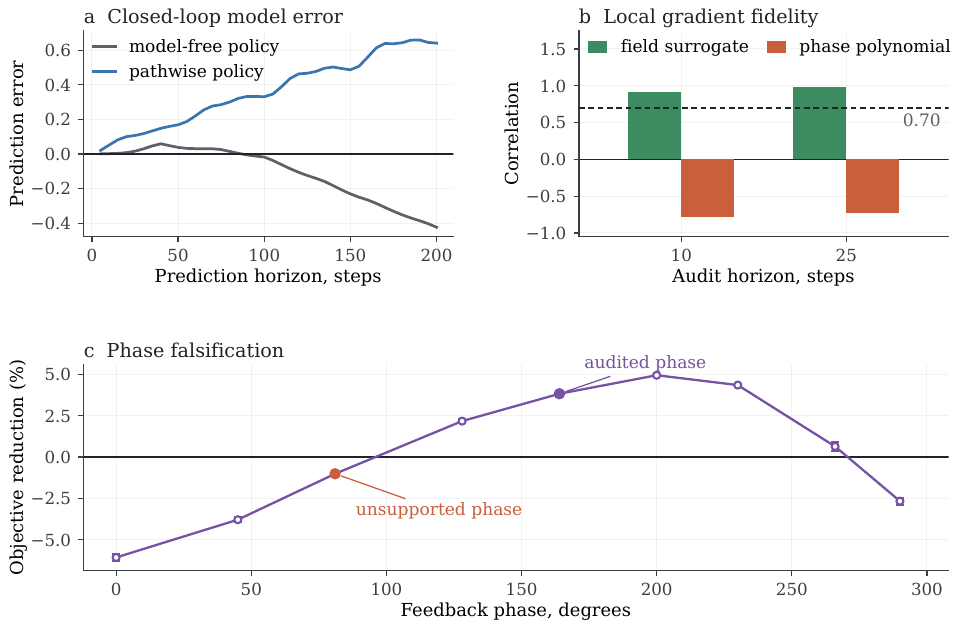}
\caption{An accurate local gradient does not prevent optimization through the same surrogate from
selecting an unsupported controller. Panel a shows the signed error of the reward predicted by the surrogate over the prediction horizon along the closed-loop trajectories of a model-free policy and of a pathwise policy. Panel b compares the field surrogate with a phase polynomial at two audit
horizons against the threshold $\tau_1=0.70$, on 18 states from the closed-loop trajectories of each surrogate's own controller, a different set from the eight-state audit of Section 4.1 of the main paper. In panel c the audited phase is the one the measured
gradient selects, and the unsupported phase is the one unregularized optimization selected.}
\label{fig:supp-cylinder-mechanism}
\end{figure}

Four separate measurements locate the source of the cylinder result, and Figure~\ref{fig:supp-cylinder-mechanism} summarizes them. A harmonic fitted to the measured gradient is stable against the loss of any single state, while the extracted phase varies by about $20^\circ$ across independent repetitions of the pipeline. Freezing the shedding period costs $0.16$ percentage points at Reynolds number 200. The anchor recovers the audited phase at every
weight, and a penalty on the local gradient sign recovers it at none.

\paragraph{Uncertainty on the extracted phase.}
The phase is fitted, so it has two sources of uncertainty. The law is fitted to the action gradient of the latent surrogate at 18 states, six on the closed-loop trajectory of each of three training seeds, and a harmonic fitted to the measured gradient at the same states gives $163^\circ$. Leaving out one state at a time moves the fitted law phase between $154^\circ$ and $171^\circ$. A harmonic fitted directly to the eight audited measurements is more stable and varies within $4.2^\circ$ under the same test. Repeating the whole
pipeline of surrogate fitting, finite-difference evaluation, and extraction at three independent seeds is far
less stable, giving $143.5^\circ$, $164.3^\circ$, and $146.0^\circ$. The measured relation
between phase and gain is steep here, $0.046$ percentage points per degree between
$128^\circ$ and $164^\circ$. At this slope, the spread of $18^\circ$ to $21^\circ$ corresponds to $0.8$ to $1.0$ percentage points, about six to seven times the standard deviation of $0.14$ over initial conditions. One
further clarification prevents a misreading of the released result. That policy is both
initialized and anchored at $164^\circ$. Its four training seeds finish between $161^\circ$ and $164^\circ$, which measures how far release moves a policy from its own target. That measure is a separate quantity from the extracted phase.

\paragraph{The frozen shedding period across Reynolds number.}
The control law estimates the shedding phase with a period fixed at $5.56$, the nominal value at Reynolds number 100, measured here as $5.59$. The measured period falls to $5.08$, $4.81$, and $4.63$ at
Reynolds numbers 150, 200, and 250. The period is held fixed because changing it would
refit part of the law. Holding it fixed distorts the estimated phase by at most $5.2^\circ$, with a root mean square of at most $3.7^\circ$ and no systematic offset. A direct comparison at Reynolds number 200 between the frozen
period and the locally measured one puts the cost at $0.16$ percentage points, $4.34\%$ against $4.50\%$.

\paragraph{Anchor weight and gradient sign sweeps.}
The anchor weight of the pathwise loss was set to 1, 10, and 100 in turn. The final
feedback phases are $161.4^\circ$, $164.6^\circ$, and $164.1^\circ$, at harmonic fit quality
$0.979$, $0.990$, and $0.990$. All three recover the audited phase, so the repair is
insensitive to this weight across two orders of magnitude. Four training seeds at the middle setting give $163.0^\circ$ to $165.1^\circ$. Decaying the
anchor from 10 to 0 or from 10 to 1 gives $164.2^\circ$ and $162.5^\circ$. All of these lie near the audited phase of $164^\circ$. The
same sweep was run for the alternative that replaces the law anchor with a penalty matching the
sign of the surrogate action gradient at the current state. Weights of 0.1, 1, and 10 give
$71.7^\circ$, $66.6^\circ$, and $66.5^\circ$, all close to the unsupported phases that optimization without the anchor selects. Therefore, local sign agreement can be high while optimization
selects a feedback phase the high-fidelity measurements do not support. These sweeps
use the frozen surrogate and one optimization seed per weight, and four at the middle weight, so they check the mechanism,
and controller performance is measured separately in the solver.

\paragraph{Reynolds number 200 model-free baseline.}
PPO was trained in the high-fidelity solver for 16,384 interactions per seed. Selecting the best
checkpoint for each of three seeds gives $7.63\pm1.43\%$, a sample standard deviation over seeds, biased upward by that selection.

\paragraph{Pathwise optimization at Reynolds number 100.}
The latent surrogate supplied a useful local action gradient field while
pathwise optimization from an unguided initialization selected an unsupported phase
region. This occurred even at a one-step horizon. Behavioral clones of the failed
pathwise policies remained at the same unsupported phase. A policy supervised directly
by the audited gradient field selected the useful phase. Phase and action amplitude were tested together, because changing the amplitude can reverse
the ordering of candidate phases. The amplitude is the bound $\alpha$ that caps the extracted law.

\section{Three cylinder cluster evidence}
\label{app:pinball}

The three cylinder cluster separates prediction accuracy from action gradient accuracy. Appendix~\ref{app:pinball-re100} explains why the study uses $Re=30$. It also gives the largest seed spread after the anchor penalty is removed.

\subsection{The inconclusive preliminary at Reynolds number 100}
\label{app:pinball-re100}

A preliminary uncontrolled calculation at $Re=100$ was run before the reported campaign, to
check whether the higher Reynolds number produces the periodic oscillation that the simulated flow lacks at $Re=30$. It
does not. On the medium mesh the tail mean objective is $4.359$ with sample standard deviation
$0.005$, a fluctuation of $0.12\%$ of the mean. On the fine mesh it is $3.909$ with sample
standard deviation $0.015$, a fluctuation of $0.39\%$. Consequently, the two meshes disagree by
$11.5\%$ in the tail mean, which is far larger than the fluctuation on either mesh, and
neither mesh shows a clear oscillation. No controller was involved. Because the objective itself is
mesh dependent at this condition, the calculation cannot support a controller comparison, and the
reported three cylinder results use $Re=30$ on the medium mesh throughout.

\subsection{Why the first surrogate was rejected}

The first nine-scalar force-state surrogate predicted the scalar objective
well, with validation correlation 0.9996. It failed the high-fidelity gradient
audit, with pooled correlation $0.459$. Reweighting the gradient loss did not
recover the missing controlled state information. This is a direct example of
prediction accuracy without usable action sensitivity.

The replacement uses the full field, a latent Transolver, and 16-step pushforward training,
in which the model is unrolled on its own predictions during fitting, and gradient-label supervision. On 72 held-out gradient labels from 55 states, 216 comparisons in all, its pooled correlation, sign agreement and median per-actuator correlation are $0.938$, $0.968$ and $0.908$. On 48 comparisons from 16 additional states they are $0.936$, $0.938$ and $0.638$.
The second audit is split by state. The first is split by gradient label, and each state carries two labels, one at zero action and one at the action of the data-collection policy. Consequently, 38 of its 55 held-out states also appear in the fitting set. For 32 of them the fitting label is at the other reference action, and for the six states from the two uncontrolled rollouts both reference actions are zero, so the held-out label repeats a fitting label.

\subsection{Extracted law from the rejected surrogate}

The audit decision changes the controller. The extracted law of Section 4.2 of the main paper is fitted to the action gradient of the accepted latent field model at 72 stored states. The same fit, with the same states, law class, ridge coefficient and target normalization, was repeated with the action gradient of the rejected force model. The resulting law was evaluated with the same gain of four on the eight held-out initial conditions, against the same uncontrolled references. It raises the objective by $18.04\pm0.01\%$, where the uncertainty is the sample standard deviation over the eight conditions, while the law fitted to the accepted surrogate reduces it by $16.04\%$. The paired difference is $-34.08$ percentage points, with a $95\%$ interval from $-34.09$ to $-34.06$ over the eight conditions. The two laws already differ before any solver evaluation. On the 16 new held-out audit states, the action of the law from the rejected surrogate has median cosine $0.24$ with the measured descent direction, against $0.98$ for the law from the accepted surrogate. The comparison was fixed in a protocol written before the law was fitted.

\subsection{Final evaluation protocol}

The final comparison uses a 600-step scored horizon and eight frozen initial
conditions. All controller comparisons use the held-out eight conditions. Earlier mechanism and protocol-sensitivity results are retained only in
Appendix~\ref{sec:moved-results}, so they are not mixed with the final result.

A gain selection fixed in advance then scales the same extracted direction.
Three selection conditions choose gain four before the eight held-out
conditions are evaluated. The selected law reduces the objective by
$16.04\pm0.01\%$ over 600 scored steps. The uncertainty is the sample
standard deviation across the eight paired reductions, which range from
$16.02$ to $16.05\%$. It exceeds the anchor-held policy of the eight-condition robustness protocol, $9.76\%$, by $6.27$
percentage points and remains $1.53$ percentage points below matched
model-free training. Every zero-action reference was frozen and hash checked before
scoring.

A further campaign, predefined in the same way, then trained pathwise controllers from that
selected law under two anchor treatments and evaluated both in the solver. Four
training seeds and the same eight initial conditions give 32 paired rollouts per
treatment, 64 in all, at 600 scored steps each. Holding the anchor gives
$17.10\pm0.26\%$, where the uncertainty is the sample standard deviation across
the four training seed means, all 32 pairs are positive, and the pairs span
$16.84$ to $17.45\%$. Decaying the anchor to zero gives
$9.18\pm2.86\%$, again a sample standard deviation across the four training seed means of the same 32 pairs, with pairs spanning $4.97$ to
$11.38\%$ and per seed means of $11.35$, $10.44$, $4.98$ and $9.95\%$. Therefore, the released controller falls below the reference value of $9.76\%$, fixed in advance, while the anchored controller exceeds it. The released controller also actuates
harder, with mean absolute summed lift $0.419$ against $0.137$ for the anchored
controller at almost the same mean requested action norm, $1.36$ against $1.34$. An
independent verification job recomputed all 64 rollouts and 38,400 scored steps
from the stored chunks and passed 544 checks. Both controllers are evaluated on held-out conditions
under the paired protocol of Equation~(4) of the main paper.

The matched model-free baseline uses proximal policy optimization with the
same observation, action limits, objective, initial conditions, and scored
horizon. Each of the three training seeds uses 120,320 solver control
interactions. Checkpoint selection uses the final checkpoint only. The mean
reductions for the three seeds are $17.54$, $17.28$, and $17.88\%$.
Their mean is $17.57\pm0.30\%$, where the uncertainty is the sample standard
deviation across training seed means. The selected gain four law recovers most
of this reduction. The anchored controller closes the remaining gap to less
than half a percentage point.

\section{Forced two-dimensional turbulence evidence}
\label{app:turbulence}

Each of the eight training seeds passes the controller check, and the controller
reduces enstrophy by about $2.7$ times the reduction the model-free baseline reaches.
The action space contains four harmonic forcing amplitudes. The prospectively
frozen cone-residual controller constrains its correction to a five degree cone
around the audited direction. The controller check accepts a trained controller when two conditions hold. The median cosine between its action and the measured descent direction is at least $0.70$, and its mean action lies within $15^\circ$ of the action of the extracted law, which points along the negative direction of the forcing mode with amplitude $a_4$. That direction lies about $3^\circ$ from the mean measured descent direction. All eight training seeds pass the check, giving a pass rate of one and an exact binomial $95\%$ interval from
$0.631$ to $1.000$. No high-fidelity evaluation diverged.

Across training seed means the controller reduces enstrophy by
$82.01\pm0.19\%$. PPO, trained with four independent seeds and scored on the same eight evaluation conditions, reduces it by $30.63\pm21.43\%$, a sample standard deviation over the four training seed means, with seed means of $26.82$, $21.21$, $61.61$ and $12.88\%$. The controller advantage is $51.38$ percentage points, with a Welch $95\%$ confidence interval from $17.28$ to $85.48$ over the training seeds of both methods, and a bootstrap interval from $30.48$ to $65.67$ that resamples both sets of seeds. PPO seed 0 was trained before the other three, and its re-evaluation reproduces the stored values exactly. The predefined performance criterion requires at least seven controllers to pass the controller check, no diverged high-fidelity evaluation,
and a positive lower bootstrap endpoint. All three conditions pass.

\subsection{Anchor release on an anchor-regularized configuration}

\begin{figure}[htbp]
\centering
\includegraphics[width=\textwidth]{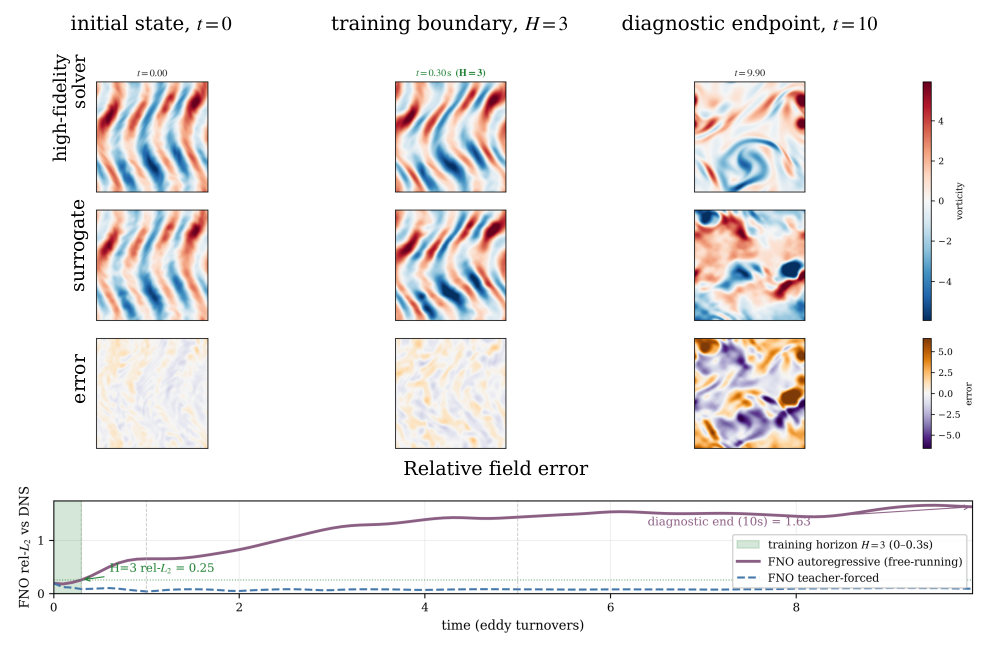}
\caption{Long-horizon spatial drift in the surrogate is larger than the drift over the horizon the
controller is trained on. The high-fidelity field, the surrogate prediction, and their error are
shown for the neural operator surrogate. In the axis
labels, FNO denotes the neural operator surrogate and DNS denotes the high-fidelity solver. Teacher-forced means the surrogate is given the true field at every step,
and free-running means it consumes its own prediction.}
\label{fig:supp-kolmogorov-spatial}
\end{figure}

The controller reported above is constrained by its architecture, which confines its action to a
five degree cone around the audited direction. Its anchor penalty weight is zero, so anchor release does not apply to it. A separate anchor-regularized configuration on
this flow isolates the effect of the anchor, under its own predefined protocol. Evaluating the policy after the last training update, under a held and under a decayed anchor schedule, from three initial conditions gives
$52.24\%$ enstrophy reduction with the anchor held and $9.48\%$ with the
anchor decayed to zero. Therefore, release retains $18.1\%$ of the gain, and one
of the three seeds falls below the uncontrolled flow. Both retrained configurations reproduced
their stored checkpoints at a maximum absolute weight difference of $0.0$. The
comparison uses training seed zero in both configurations, so it is paired and has no seed
spread, and neither final actor is offered as a controller for this flow. Figure~\ref{fig:supp-kolmogorov-spatial} shows how the prediction of the neural operator surrogate drifts from the high-fidelity field over time.

\section{Parametric chaotic equation evidence}
\label{app:chaotic}

At $\mu=0.225$ the gradient field at the fixed reference action has state-independent fraction $f=0.986$ in Table~\ref{tab:feedback-content}. That value does not describe the field the controllers meet in deployment at $\mu=0.121$. A
further audit measures high-fidelity gradients around the exact actions visited by
three deployed controllers. Table~\ref{tab:supp-ks-deployed-audit} reports the result. Around these actions the state-independent fraction is $0.024$ for the transferred controller and $0.028$ and $0.029$ for the two controllers built at $\mu=0.121$, so the measured gradient is strongly state dependent. The comparison with $0.986$ at the fixed reference action at $\mu=0.225$ also spans the change of forcing amplitude.

\begin{table}[htbp]
\centering
\caption{The gradient model of each deployed controller fails all three acceptance criteria when the derivative is measured around the actions of that controller. The audit triples and state-independent fractions are
measured at
$\mu=0.121$. Each audit triple is pooled correlation, sign agreement, and median
per-actuator correlation. All three miss the predefined thresholds $0.70$, $0.70$, and $0.50$.}
\label{tab:supp-ks-deployed-audit}
\begin{tabular}{lcc}
\toprule
Deployed controller & Audit triple & $f$ \\
\midrule
First controller built at $\mu=0.121$, $r=0.80$ & $0.049/0.500/{-}0.306$ & $0.028$ \\
Second controller built at $\mu=0.121$, $r=0.50$ & ${-}0.018/0.438/{-}0.016$ & $0.029$ \\
Controller transferred from $\mu=0.225$ & ${-}0.005/0.495/{-}0.093$ & $0.024$ \\
\bottomrule
\end{tabular}
\end{table}

The two controllers built at $\mu=0.121$ keep the base policies and the correction rule of the $\mu=0.225$ controller, replace its gradient model by one fitted to new finite differences at $\mu=0.121$, and use the strengths $r=0.80$ and $r=0.50$. The refitted gradient model passes its audit at the fixed reference action and fails around the actions of these controllers, at the same amplitude, so the change of action alone causes the failure. On eight held-out initial fields the two controllers reach accumulated state costs of $493.99$ and $471.05$, that is, $6.8$ and $6.5$ times the cost of matched TD3, $72.70$, on the same fields.

These measurements falsify the proposed explanation that the transferred controller
succeeds by retaining the fixed component seen at the reference action. Regressions that predict the measured gradient from instantaneous sensor values, from full fields or from sensor histories also fail the acceptance criteria when each source trajectory is held out in turn, with pooled correlation $-0.188$, sign agreement $0.482$ and median per-actuator correlation $-0.180$ overall. Therefore, the deployed-action measurements leave the mechanism unresolved.

A separate predefined experiment then identifies the benchmark sensor-noise floor
as the supported mechanism. The controlled sensor signal before noise is added is $0.959$ times the
noise standard deviation at $\mu=0.121$, compared with $2.05$ at $\mu=0.225$.
Adding benchmark noise at the lower amplitude raises transferred-controller cost by
$71.06$, or $109.1\%$. This result motivates the controller with observer evaluated
below.

At $\mu=0.225$, the surrogate-based controller costs $361.58\pm8.31$, a sample standard deviation over eight
held-out fields. Matched plain TD3 costs $409.56$, and matched polynomial TD3
costs $538.71$. The paired interval against plain TD3 overlaps zero, while the
surrogate-based controller has lower cost than polynomial TD3 on every field.

The controller deployed at $\mu=0.121$ is the $\mu=0.225$ controller with the strength $r$ of its last transformation, $(1-r)a+0.999\,r\,\mathrm{sign}(a)$ for each component $a$, chosen again at $\mu=0.121$. A comparison of the strengths $0.60$, $0.70$, $0.80$, $0.90$ and $0.999$ and the earlier strength $0.50$ on two initial fields ranked $0.999$ first, with a mean cost of $133.51$ against $155.82$ for $0.90$. A validation on four further fields then kept $0.999$ against the controller with strength $0.50$ and an earlier controller that applies a correction of fixed size from the same gradient model. All these fields are disjoint from the evaluation fields. The protocol fixed in advance for the transfer had fixed $0.90$, so this choice departs from it. At $\mu=0.121$, the controller with observer costs $74.92\pm4.01$, a sample standard deviation over eight
held-out fields. Matched plain TD3 costs $71.93$, and matched polynomial TD3
costs $83.75$. The paired observer minus plain TD3 difference is $+2.99$, with
a $95\%$ interval from $-1.09$ to $+7.08$. Its upper limit lies below the fixed
$10\%$ noninferiority margin of $7.19$. The observer minus polynomial TD3
difference is $-8.83$, with an interval from $-11.74$ to $-5.91$.

\begin{table}[htbp]
\centering
\caption{At $\mu=0.225$ the comparison detects no difference between the surrogate-based controller and matched TD3, and at $\mu=0.121$ the controller with observer is noninferior to matched TD3. Uncertainty is the sample standard deviation across eight initial fields. Each paired comparison is computed field by field against the controller named in the last column, and a positive paired difference means the row controller has the higher cost. Costs at the two forcing amplitudes cannot be compared with each other. Matched TD3 and polynomial TD3 are single trained checkpoints, and their spread across fields is not reported.}
\label{tab:ks-scores}
\small
\begin{tabularx}{\textwidth}{>{\raggedright\arraybackslash}p{0.30\textwidth}
>{\raggedleft\arraybackslash}p{0.16\textwidth}>{\raggedright\arraybackslash}X}
\toprule
Controller & Accumulated state cost & Paired comparison, $95\%$ interval \\
\midrule
\multicolumn{3}{l}{\textit{Forcing amplitude $\mu=0.225$}} \\
surrogate-based controller & $361.58\pm8.31$ & $-47.98$ from $-137.04$ to $+41.09$ against matched TD3, no detected difference \\
matched TD3 & $409.56$ & baseline \\
matched polynomial TD3 & $538.71$ & $+177.13$ from $+152.46$ to $+201.81$ against the surrogate-based controller \\
\addlinespace
\multicolumn{3}{l}{\textit{Forcing amplitude $\mu=0.121$}} \\
surrogate-based controller with observer & $74.92\pm4.01$ & $+2.99$ from $-1.09$ to $+7.08$ against matched TD3, noninferior under the $7.19$ margin fixed in advance \\
matched TD3 & $71.93$ & baseline \\
matched polynomial TD3 & $83.75$ & $+8.83$ from $+5.91$ to $+11.74$ against the surrogate-based controller with observer \\
\bottomrule
\end{tabularx}
\end{table}

\section{Per-flow result figures}
\label{app:per-flow-figures}

\begin{figure}[htbp]
\centering
\includegraphics[width=\textwidth]{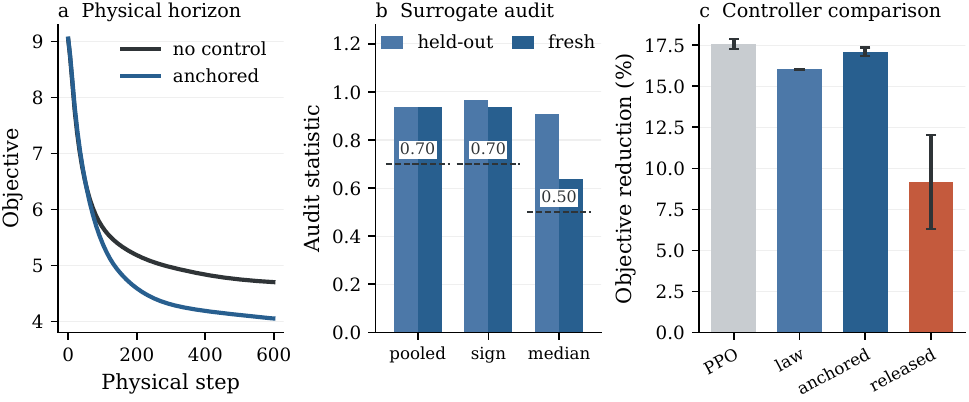}
\caption{Anchor regularization recovers most of the reduction that model-free training reaches,
and removing it loses about half of that. Panel a compares the uncontrolled objective with the objective under the
policy that holds the anchor, over one 600-step horizon. Panel b shows the two gradient audits against the acceptance thresholds, $\tau_1=\tau_2=0.70$ for the correlation and the sign agreement and $\tau_3=0.50$ for the median per-actuator correlation, on
gradient labels held out of surrogate fitting and on 16 new states. Panel c compares matched
model-free training with the extracted law, the policy that holds the anchor, and the policy after
the anchor is removed. Its rows are PPO over three training seeds, the extracted law over eight
initial conditions, and the anchored and released policies over four training seeds each. Error bars
show sample standard deviations across training seed means for the model-free, anchored and released rows, and across the eight initial conditions for the extracted law.}
\label{supp-fig:pinball-overview}
\end{figure}

\begin{figure}[htbp]
\centering
\includegraphics[width=0.94\textwidth]{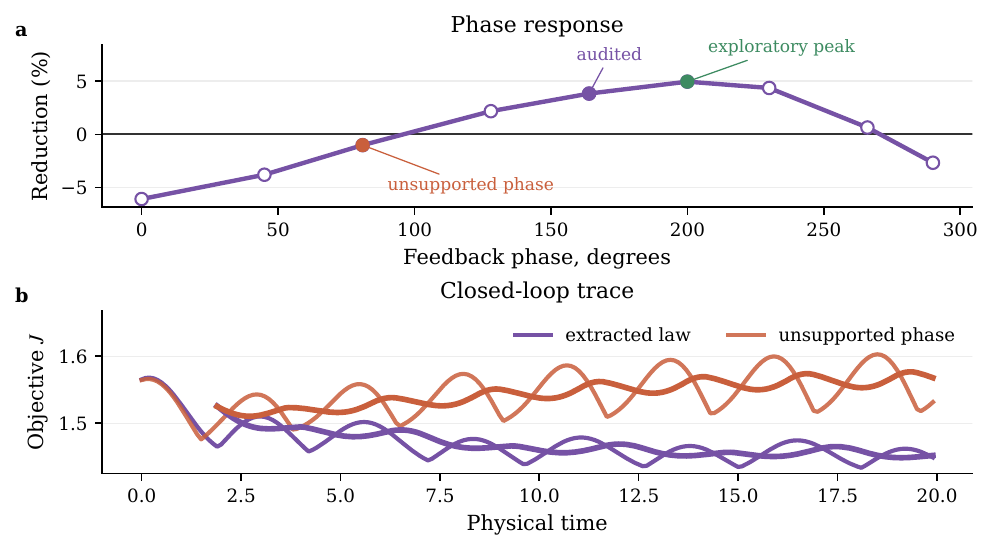}
\caption{Objective reduction varies smoothly with the feedback phase, and the audited phase lies
near the measured optimum. Panel a shows paired
high-fidelity reduction as the feedback phase changes at fixed action amplitude.
The $164^\circ$ law and the $81^\circ$ unsupported phase were fixed before
evaluation. The full curve and its maximum at $200^\circ$ are exploratory.
The standard deviations over eight paired initial conditions, at most $0.25$ percentage points, are smaller than the markers.
Panel b shows representative objective histories. Thick curves are running
means over 20 control steps. Time samples are not independent replicates.}
\label{fig:supp-distilled-controller}
\end{figure}

\begin{figure}[htbp]
\centering
\includegraphics[width=\textwidth]{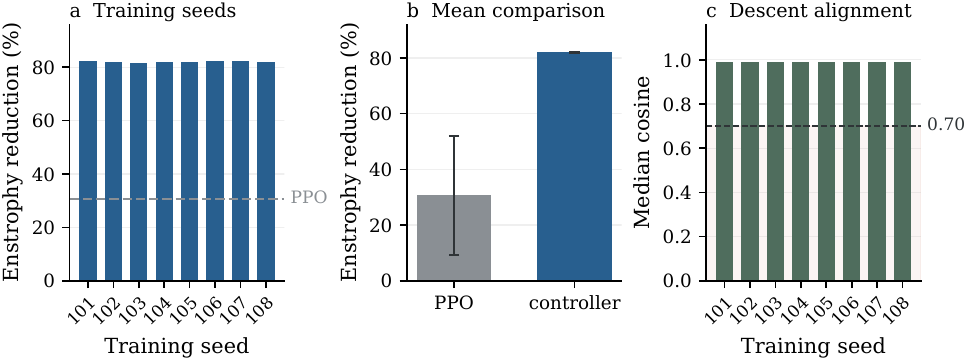}
\caption{All eight training seeds, fixed before training, pass the controller check and exceed the model-free baseline. Panel a shows the enstrophy reduction for
each of eight controller training seeds. Every seed passes the controller check, and the dashed line is the mean of four independently trained PPO seeds. Panel b compares the mean across training seeds with PPO. The error bar is the sample standard
deviation across the eight training seed means, and for PPO across its four training seed means. The mean advantage over PPO is $51.38$ percentage points, with a Welch $95\%$ confidence interval from $17.28$ to $85.48$. Panel c shows the median cosine between the controller action and the
high-fidelity descent direction for every training seed, with the threshold of $0.70$ of the controller check. Each
training seed is evaluated over eight held-out initial conditions.}
\label{fig:supp-kolmogorov-overview}
\end{figure}

\begin{figure}[htbp]
\centering
\includegraphics[width=\textwidth]{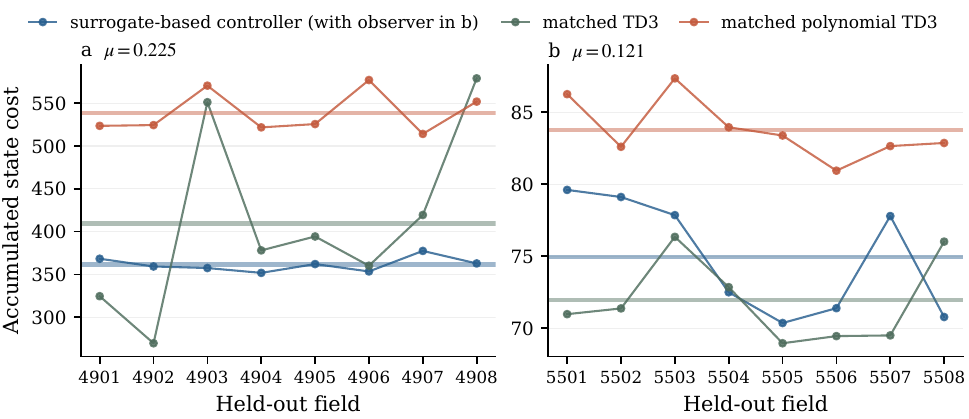}
\caption{The comparison detects no difference from plain TD3 at $\mu=0.225$, the controller with observer is noninferior to it at $\mu=0.121$, and both cost less than polynomial TD3, on eight held-out fields each. Panel a compares the surrogate-based controller with matched plain and polynomial TD3 at
$\mu=0.225$. Panel b makes the same field-matched comparison for the controller with observer at $\mu=0.121$. Thin lines show individual fields and heavier horizontal
lines show controller means. Lower accumulated state cost is better.}
\label{fig:supp-ks-overview}
\end{figure}

The figures below, Figures~\ref{supp-fig:ladder} and~\ref{fig:supp-cylinder-all-arms} for the rotating cylinder, Figure~\ref{fig:supp-distilled-controller} for the phase response of the extracted law, Figure~\ref{supp-fig:pinball-overview} for the three cylinder cluster, Figure~\ref{fig:supp-kolmogorov-overview} for forced turbulence and Figure~\ref{fig:supp-ks-overview} for the chaotic equation, show, for each test case, the per-seed and per-condition scores that the main
text summarizes as a mean and a standard deviation.

\begin{figure}[htbp]
\centering
\includegraphics[width=\textwidth]{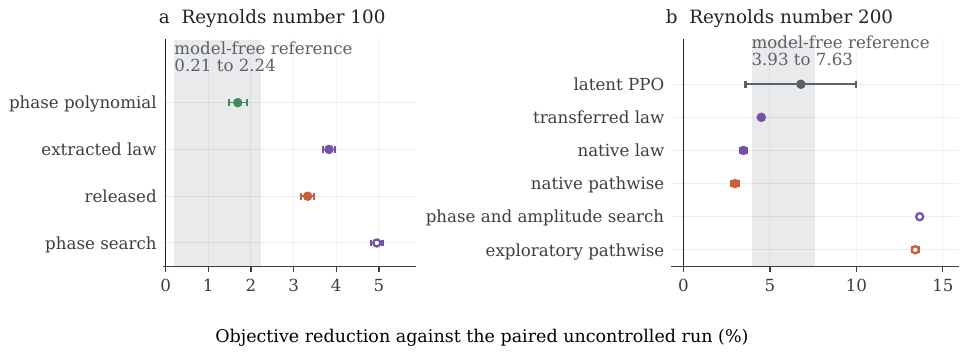}
\caption{The extracted law exceeds every trained controller at Reynolds number 100, and at Reynolds number 200 the model-free baseline exceeds the native law and the native pathwise controller, and only the exploratory rows exceed the model-free baseline. The transferred law at Reynolds number 200 uses the shedding period measured at that Reynolds number, which gives $4.50\%$ against $4.34\%$ with the period of Reynolds number 100. The left panel shows
Reynolds number 100. The right panel shows Reynolds number 200. Error bars are
standard deviations over training seeds or paired initial conditions as stated
in the text. The shaded band spans the two model-free PPO references, the fixed budget checkpoint and the selected peak checkpoint. At Reynolds number 100 these are $0.21$ and $2.24\%$, and at Reynolds number 200 they are $3.93$ and $7.63\%$. The latent PPO row is model-based and is not part of that band, and it is the best training iteration scored against the mean uncontrolled objective. PPO uses 32,768 and 16,384 solver interactions per seed. The model-based training configurations use about 4,600. Hollow markers denote results that inherit an exploratory high-fidelity search.}
\label{supp-fig:ladder}
\end{figure}

\begin{figure}[htbp]
\centering
\includegraphics[width=\textwidth]{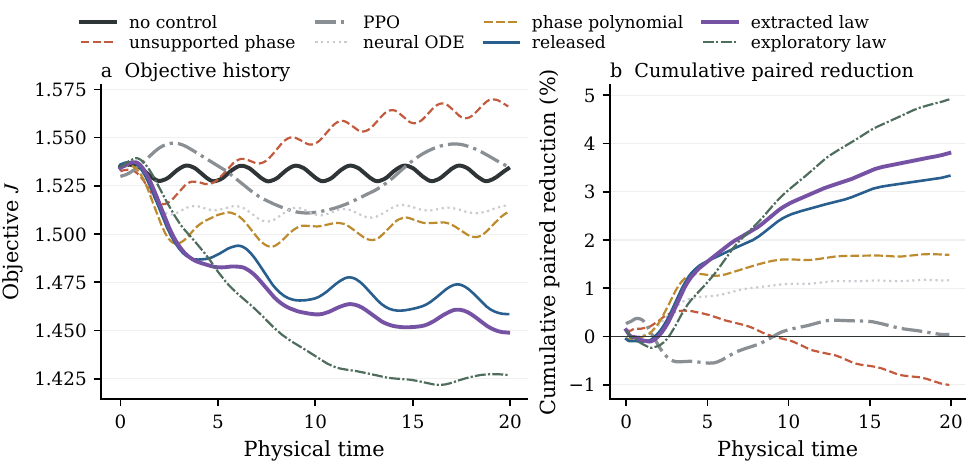}
\caption{The extracted law separates from every trained controller within the first 50 control
steps and holds that separation to the end of the horizon, at Reynolds number 100. Panel a shows the objective under the uncontrolled flow and under seven controllers, including the phase polynomial and neural ordinary differential equation pathwise controllers. Each curve is a mean over the paired restart states available to that controller, smoothed over 15
control steps. The uncontrolled objective, PPO, and the extracted law are drawn heavier
because they carry the comparison, and the remaining diagnostics are drawn lighter. Panel b divides the running
paired reduction by the running uncontrolled objective, so each curve ends near the
full horizon reduction reported for that controller, except PPO, whose 300-step rollouts from four other restart states are truncated here at 200 steps. The model-free baseline draws from four restart states, while every other controller draws
from the same bank of eight. Therefore, it is matched in objective and solver, and unmatched in
initial state.}
\label{fig:supp-cylinder-all-arms}
\end{figure}

\section{Simulation cost and reproducibility}
\label{app:cost-repro}

The surrogate-based route costs fewer high-fidelity solver steps than one model-free training run
on every test case. Every count below is read from the stored result files of the corresponding runs. Table~\ref{tab:supp-fd-cost} lists the rotating cylinder finite-difference campaigns, and Table~\ref{tab:supp-cost} collects the counts for every test case.
The Firedrake and chaotic-equation evaluations run on CPU-only allocations, and the forced-turbulence solver runs on one GPU. Firedrake three cylinder jobs
use two CPUs and are serialized. 

The model-free cylinder baseline uses 32,768 solver interactions at $Re=100$ and
16,384 at $Re=200$. Cross-method ratios require a complete record of solver
interactions. For the Reynolds number 100 phase polynomial study, each model-based
training seed uses 3,000 offline transitions and eight on policy collections of
200 transitions. The total is 4,600. Each model-free PPO seed uses 32,768
training interactions. The paired gradient development data use 1,500 solver
branch interactions. The independent finite-difference campaign uses 3,600. The
predefined extracted law uses 1,600 evaluation interactions. The released
pathwise controller uses 6,400 evaluation interactions. The larger exploratory
law sweep uses 17,600 evaluation interactions. 

\subsection{Finite-difference cost and the efficiency denominator}

Four conditions set how these figures are read. Solver steps advancing to each
finite-difference restart state are not included, so every finite-difference total is a lower
bound. The accounting counts high-fidelity solver steps alone, which is the resource the method
exists to save, so surrogate training is not included in any column. The forced-turbulence
column covers the audit and the controller. Solver families differ across test cases, so no cost
is summed across them and no single cross-flow efficiency figure is claimed. The campaign-by-campaign breakdown on the rotating cylinder follows below.

Each stored cylinder campaign holds $n_{\rm states}$ high-fidelity restart states. From
each state the solver is branched once per entry of the action grid and advanced
$H$ control steps, so the cost is $n_{\rm states}\,n_{\rm actions}\,H$.

\begin{table}[htbp]
\centering
\caption{The finite-difference campaigns behind the reported result cost 2,766 control steps under
the distinct-state count. The campaigns are measured at $Re=100$. Rows are the stored entries and
distinct states are the unique starting flow states among them. Step counts use distinct states,
and every campaign listed uses the lift-inclusive objective.}
\small
\begin{tabular}{lrrrrrl}
\toprule
Finite-difference campaign & Rows & Distinct & Actions & $H$ & Steps & Role \\
\midrule
paired actions, horizon 25 & 12 & 12 & 5 & 25 & 1,500 & supporting \\
lift resolution, horizon 25 & 16 & 4 & 9 & 25 & 900 & supporting \\
converged response, horizon 25 & 4 & 2 & 3 & 25 & 150 & supporting \\
lift resolution, horizon 5 & 16 & 4 & 9 & 5 & 180 & resolution check \\
lift resolution, horizon 1 & 16 & 4 & 9 & 1 & 36 & resolution check \\
\bottomrule
\end{tabular}
\label{tab:supp-fd-cost}
\end{table}

\begin{table}[htbp]
\centering
\caption{The complete audited route costs fewer high-fidelity solver steps than one model-free
training run on every test case. Every entry is a count of solver control steps read from the stored result files of the runs. The two rotating cylinder values count the same five finite-difference campaigns, the first by distinct
restart states and the second by stored rows, because several campaigns hold up to four copies of
each state. The rotating cylinder and the three cylinder cluster report a model-free reference of three training seeds, forced two-dimensional turbulence reports four, and the parametric chaotic equation reports one, so for the parametric chaotic equation the last two advantage rows coincide. The forced-turbulence column covers the audit and the controller. The three cylinder column counts the field data set, the 144 fitting and held-out gradient labels and the second audit, and leaves out the 4,680 solver steps that selected the gain. The $\mu=0.225$ column counts the 44,800 finite-difference steps that label the gradient model and the 47,952 steps of the campaign that selected and evaluated the deployed strength, and leaves out the earlier development campaigns and the second audit of the gradient model.
The $\mu=0.121$ column charges the runs that chose the strength of the last transformation at that amplitude, $23{,}976$ interactions plus $48$ for the preliminary check, and the development and held-out evaluation of the observer, $39{,}960$. The controller is transferred from $\mu=0.225$, so neither its construction nor its finite differences appear here, and the finite-difference entries are zero. The deployed-action audit of Table~\ref{tab:supp-ks-deployed-audit} is a separate diagnostic outside this route, with a stored upper bound of $72{,}585$ solver interactions.}
\label{tab:supp-cost}
\footnotesize
\begin{tabularx}{\textwidth}{>{\raggedright\arraybackslash}p{0.235\textwidth}
>{\raggedleft\arraybackslash}p{0.115\textwidth}>{\raggedleft\arraybackslash}p{0.095\textwidth}
>{\raggedleft\arraybackslash}p{0.095\textwidth}>{\raggedleft\arraybackslash}p{0.105\textwidth}
>{\raggedleft\arraybackslash}X}
\toprule
Route or baseline & Rotating cylinder & Forced turbulence & Three cylinder cluster & Chaotic $\mu=0.225$ & Chaotic $\mu=0.121$ \\
\midrule
finite differences alone & 2,766 to 6,264 & 12,800 & 1,330 & 44,800 & 0 \\
route through the extracted law & 5,766 to 9,264 & 14,400 & 4,930 & 92,752 & 0 \\
complete route & 7,366 to 10,864 & 14,400 & 4,930 & 92,752 & 63,984 \\
\addlinespace
one model-free training run & 32,768 & 100,000 & 120,320 & 9,990,000 & 9,990,000 \\
model-free baseline as reported & 98,304 & 400,000 & 360,960 & 9,990,000 & 9,990,000 \\
\addlinespace
advantage over one training run & $3.0$ to $4.4$ times & $6.9$ times & $24$ times & $108$ times & $156$ times \\
advantage over the reported reference & $9.0$ to $13.3$ times & $28$ times & $73$ times & $108$ times & $156$ times \\
finite-difference share of the audited route & $38$ to $58\%$ & $89\%$ & $27\%$ & $48\%$ & $0\%$ \\
\bottomrule
\end{tabularx}
\end{table}

Several campaigns store up to four identical copies of each starting state. Counting distinct
states gives 2,766 control steps for the finite differences behind the reported result, and
counting stored rows gives 6,264. Both counts are reported so the ratio can be recomputed under
either.

Every comparison is given with the quantity it is divided by. Under the distinct-state reading,
the extracted law route costs 5,766 control steps and the pathwise route costs 7,366. Under the
conservative row reading, they cost 9,264 and 10,864. One model-free training run costs 32,768,
and the three run reference costs 98,304. Therefore, the audited pathwise approach costs 3.0 to 4.4
times fewer control steps than one training run, and 9.0 to 13.3 times fewer than the three run
reference.
The finite-difference share is 37.6 to $57.7\%$ of the pathwise route.

The finite-difference error is not estimated in any test case, because no stored campaign repeats a solve independently. The turbulence and
chaotic-equation solvers are deterministic, and the cylinder campaigns hold only identical copies
of each restart state.

The cylinder finite-difference sets at horizons 1, 5, and 25 share only four distinct starting
states, so no agreement statistic across horizons has a useful sample. Over those four states the
gradient magnitude rises from $0.0020$ at horizon 1 to $0.0267$ at horizon 25. Therefore, whether a shorter finite difference recovers the same field is not settled here.

The three cylinder finite difference is a one-step central difference costing $2n_a+1=7$
control steps per gradient label, six perturbed branches and one base step. The 144 fitting and held-out gradient labels of the field data set cost 1,008 finite-difference steps. The second held-out audit uses 16
states, or 112 finite-difference steps, plus 210 orbit-advance steps to reach them. The finite differences are $27\%$ of the route, below the 38 to $58\%$ on the rotating cylinder. The difference is the finite-difference horizon,
1 against 25. Finite-difference cost scales as action dimension times finite-difference horizon. That
product is the design variable that determines whether an audit is affordable.

The forced turbulence archive records paired gradients on 20 states with four action
modes, and every cached state file stores the per-branch rollout horizon of 20 control steps.
Those counts give $2\times20\times4\times20=3{,}200$ finite-difference steps for each of the four stored branches, 12,800 in all, and 1,600 further steps for the uncontrolled rollout that each branch is compared against. The chaotic equation labels give $56\times2\times8\times50=44{,}800$ steps, for 56 labeled states, eight actuators and a horizon of 50. In both, the solver cost per step is far below Firedrake, so the audit is not the
binding constraint on either.

\subsection{Settings of the model-free baselines}
\label{sup:baseline-settings}

All model-free baselines use standard settings of the published implementations, and no hyperparameter was tuned on the evaluation conditions. Table~\ref{tab:baseline-settings} lists them.

\begin{table}[htbp]
\centering
\caption{The model-free baselines use standard settings, listed per test case. PPO uses Stable-Baselines3, and its unlisted settings are the library defaults. TD3 is the implementation of the published benchmark.}
\label{tab:baseline-settings}
\small
\begin{tabularx}{\textwidth}{>{\raggedright\arraybackslash}p{0.22\textwidth}>{\raggedright\arraybackslash}X>{\raggedright\arraybackslash}p{0.2\textwidth}}
\toprule
Test case and algorithm & Settings & Training seeds and interactions \\
\midrule
rotating cylinder, PPO & multilayer perceptron policy, learning rate $3\times10^{-4}$, 512 steps per update, batch size 128, discount $0.99$, normalized observations & three seeds, 32,768 each (16,384 at $Re=200$) \\
three cylinder cluster, PPO & multilayer perceptron policy, learning rate $3\times10^{-4}$, 512 steps per update, batch size 128, discount $0.99$, normalized observations and rewards & three seeds, 120,320 each \\
forced turbulence, PPO & convolutional policy on the vorticity field, learning rate $3\times10^{-4}$, 2048 steps per update, batch size 64, 10 epochs, discount $0.99$, generalized advantage parameter $0.95$, clip range $0.2$, entropy coefficient $0.01$ & four seeds, about 100,000 each \\
parametric chaotic equation, TD3 & actor and critic with hidden width 256, learning rate $3\times10^{-4}$, batch size 256, target update rate $0.005$, target policy noise $0.2$ clipped at $0.3$, policy update every two critic updates, exploration noise $0.1$, random actions in the first 500 episodes & one seed, 9,990,000 \\
\bottomrule
\end{tabularx}
\end{table}

\subsection{Sensor noise robustness of the extracted cylinder law}
\label{sup:noise}

The extracted law keeps its gain under sensor noise up to $59\%$ of the natural scale of the
signal it reads. It reads a single noisy measurement, which makes it the appropriate place to test
degraded sensing.
Gaussian noise was added to the lift signal that feeds the control law, at standard
deviations of $0$, $0.02$, $0.05$, and $0.10$. The lift standard deviation across the
finite-difference states is about $0.17$, so these are roughly $0$, $12$, $29$, and $59\%$ of
the natural scale of the measured signal. The objective is scored on the true lift and
drag at every level, so the noise degrades what the controller knows and never the
physics or the score. The law is frozen at $164^\circ$ and amplitude $0.2$, with no refitting and no retuning. Each level uses the same eight paired initial conditions
and the same horizon of 200 control steps as every other cylinder controller except PPO.

The gains are $3.83\pm0.14$, $3.83\pm0.18$, $3.76\pm0.19$ and $3.63\pm0.19\%$, each a sample standard deviation over the eight paired initial conditions. The protocol
required the noiseless case to reproduce the extracted-law gain of $3.83\%$ before any other row could be read, and it does so to the second decimal. Losing the
equivalent of $59\%$ of the natural lift scale to noise costs $0.20$ percentage
points of gain, which is about $5\%$ of the effect. The gain keeps its sign at
every tested level, so there is no robustness boundary inside this range. One noise seed
was used per level and initial condition, so variation across noise seeds is not
measured. This limits the scope of the test.

\subsection{Parametric chaotic equation evaluation protocol}
\label{sup:ks-protocol}

Each final comparison uses eight held-out fields that were not used for controller selection.
Controllers and matched baselines receive the same initial field and observation noise
within each source. Every rollout has 999 control interactions, and the reported score is
accumulated state cost $c_1$. Lower values are better. At $\mu=0.121$, sources, the numbered random seeds that generate an initial field and its noise sequence, 5501
through 5508 form the held-out confirmation set and the observation noise standard
deviation is $0.25$. The observer, its update weight, the TD3 checkpoints, and the
noninferiority margin are frozen before these sources are evaluated.

\subsection{Phase frame convention for the rotating cylinder}
\label{sup:phase-frame}

Every phase quoted for this flow, except the window noted at the end of this paragraph, uses the control law frame, in which the shedding phase is
$\varphi_t=\operatorname{atan2}(\dot{\widetilde C}_{L,t}/\omega_{\rm shed},
\widetilde C_{L,t})$ with $\omega_{\rm shed}=2\pi/T_{\rm shed}$ and $\widetilde C_L$ the
filtered lift. The high-fidelity measurement records phase differently, as
$\operatorname{atan2}(\dot C_{L},C_L)$ on unfiltered lift, so the two frames must never be mixed. The low-pass filter on the lift signal has time constant $0.556$. At
$\omega_{\rm shed}=1.130$ this shifts the two definitions by a fixed $28.3^\circ$, measured by
driving the filter with a sinusoid at the shedding frequency. The normalization by $\omega_{\rm shed}$ contributes
about $0.1^\circ$ on average, because its effect is antisymmetric over the cycle. Any
comparison between a surrogate phase and a finite-difference phase must be made in one declared frame. The window of $124^\circ$ to $163^\circ$ quoted for the measured gradient comes from an earlier phase-resolved analysis whose frame was not stored with it, so it can be compared with the law phase only approximately.

\subsection{Model-free baseline training curve on the rotating cylinder}
\label{sup:ppo-curve}

The three PPO training seeds were scored at 8,192, 16,384 and 32,768 solver interactions with the paired rule of Equation~(4) of the main paper over the full evaluation window. Seed 0 gives $3.11$, $2.29$ and $2.83\%$ at the three budgets, seed 1 gives $0.82$, $0.97$ and $2.08\%$, and seed 2 gives $1.53$, $-5.19$ and $-4.30\%$. The mean over seeds is $1.82\%$ at 8,192, $-0.64\%$ at 16,384 and $0.21\%$ at 32,768, so training had not converged within its budget, and the third seed ends below the uncontrolled flow. Selecting each seed's best checkpoint gives $2.24\pm0.80\%$, where the uncertainty is the sample standard deviation over the three seeds. This selection uses the evaluation results to choose the checkpoint, so it is not a held-out score, and it remains below the $3.83\%$ of the extracted law. With the training seed as the unit, the Welch test of the extracted law against the three final PPO seed means gives a difference of $3.62$ percentage points with a $95\%$ interval from $-6.10$ to $13.35$. The test of the released controller gives $3.12$ percentage points with an interval from $-6.61$ to $12.85$. Neither difference is statistically significant at three model-free seeds. Rescoring every cylinder controller with the single paired rule of Equation~(4) of the main paper moves the harmonic laws by at most $0.04$ percentage points. It gives the fixed budget model-free baseline as $0.21\pm3.54\%$, a sample standard deviation over 12 paired rollouts, and leaves the ordering of the controllers unchanged.

The released pathwise policy at Reynolds number 100 is both initialized and anchored at
$164^\circ$, and its four training seeds finish between $161^\circ$ and $164^\circ$. That
observation measures how far release moves a policy from its own target, and it is not a
second measurement of the extracted phase. At Reynolds number 200 the corresponding seeds
converge from $222^\circ$ to $224^\circ$ near amplitude $0.4$. The law estimates the shedding
phase with a period fixed at its Reynolds number 100 value, which costs $0.16$
percentage points at Reynolds number 200.

\section{Controller scores across the test cases}
\label{app:moved-results}

This appendix reports the measurements behind the controller the main text names for each test
case, together with the comparisons that set those numbers in context. Table~\ref{tab:supp-controller-scores} collects the scores. The main text reports one
controller for each test case with the baseline that gives that number its meaning. Every row below follows the same predefined protocols and is computed from the same stored results.

\begin{table}[htbp]
\centering
\caption{Releasing the anchor lowers the reduction on the three cylinder cluster, and the rotating cylinder controller optimized without the anchor penalty raises the objective.
The score is the reduction of that test case's
objective against a paired uncontrolled run started from the same state, in \%, and higher is better. The row optimized without the anchor is the exception, and its note states its reference. The scored unit is stated for each row, because some controllers are scored over initial
conditions and others over training seeds. Uncertainty is the sample standard deviation across
those units.}
\label{tab:supp-controller-scores}
\small
\begin{tabularx}{\textwidth}{>{\raggedright\arraybackslash}p{0.28\textwidth}
>{\raggedright\arraybackslash}p{0.21\textwidth}
>{\raggedleft\arraybackslash}p{0.14\textwidth}>{\raggedright\arraybackslash}X}
\toprule
Controller & Scored over & Reduction & Note \\
\midrule
\multicolumn{4}{l}{\textit{Rotating cylinder wake, $Re=100$, objective
$\overline{C_D}+0.2\,\overline{|C_L|}$}} \\
extracted law & eight initial conditions & $3.83\pm0.14$ & the audited direction, no policy training \\
anchored and released pathwise & 32 paired rollouts & $3.33\pm0.15$ & retains $87\%$ of the reduction of the extracted law \\
pathwise without the anchor & three training seeds & $-0.52\pm0.25$ & raises the objective despite an accurate gradient, with no anchor penalty and an initialization not derived from the law, best of eight training iterations per seed, each scored by one rollout against the mean uncontrolled objective \\
model-free baseline & 12 paired rollouts & $0.21\pm3.54$ & 32,768 solver interactions per
training seed \\
extracted law frozen, at $Re=150$, $200$ and $250$ & eight initial conditions each &
$4.39\pm0.10$, $4.34\pm0.15$, $4.51\pm0.22$ & transferred with no refitting of any kind \\
\addlinespace
\multicolumn{4}{l}{\textit{Three cylinder cluster, $Re=30$, objective
$\sum_j C_{D,j}+0.2|\sum_j C_{L,j}|$}} \\
extracted law & eight initial conditions & $16.04\pm0.01$ & every condition between $16.02$ and
$16.05$ \\
anchored pathwise & four training seeds & $17.10\pm0.26$ & all 32 paired rollouts positive \\
released pathwise & four training seeds & $9.18\pm2.86$ & seed means from $4.98$ to $11.35$ \\
model-free baseline & three training seeds & $17.57\pm0.30$ & 120,320 solver interactions per seed \\
\addlinespace
\multicolumn{4}{l}{\textit{Forced two-dimensional turbulence, $Re=200$, objective mean enstrophy}} \\
cone-residual controller & eight training seeds & $82.01\pm0.19$ & all eight seeds pass the controller check \\
independently trained PPO & four training seeds & $30.63\pm21.43$ & controller advantage $51.38$ percentage points, $95\%$ interval $17.28$ to $85.48$ \\

\bottomrule
\end{tabularx}
\end{table}

\label{sec:moved-results}

\subsection{Rotating cylinder wake}

Pathwise optimization through the phase polynomial surrogate gains $1.69\pm0.21\%$, a sample standard deviation over 24 paired rollouts from three training seeds. The more predictive latent surrogate gives $-0.52\pm0.25\%$ over three training seeds and so makes the flow worse, where each seed value is the best of eight training iterations, each scored by one rollout against the mean uncontrolled objective, and the final iterations give $-1.25\pm0.52\%$. Its horizon 25 action gradient nonetheless correlates at $0.996$ with the eight audited measurements, and on the 18 states visited by its own controllers its fitted phase lies $0.8^\circ$ from the measured one. A neural ordinary differential equation surrogate gives $1.16\pm0.98\%$ under the same protocol.
The extracted law at the audited $164^\circ$ phase gives $3.83\pm0.14\%$ and the
unsupported $81^\circ$ phase, fixed before evaluation as its counterpart, gives $-1.01\pm0.18\%$.

Pathwise training through the surrogate, even at a one-step horizon, selects phases from
$77^\circ$ to $122^\circ$, whereas the paired high-fidelity measurements support a window from
$124^\circ$ to $163^\circ$. Repeating surrogate fitting, finite-difference evaluation, and extraction at three seeds
gives $143.5^\circ$, $164.3^\circ$, and $146.0^\circ$, a spread worth $0.8$ to $1.0$ percentage points. A nine phase exploratory sweep at the same amplitude peaks at $4.95\pm0.13\%$ near $200^\circ$, and a high-fidelity search finds $13.66\pm0.06\%$ at Reynolds number 200.
Therefore, the frozen law captures $77\%$ of the best reduction found at the lower Reynolds number and $32\%$ at the higher one. Varying the anchor weight over two orders of magnitude recovers the audited phase every time, at $161.4^\circ$, $164.6^\circ$, and $164.1^\circ$. Penalizing the local gradient
sign instead fails at every weight and converges near $67^\circ$.

\subsection{Three cylinder cluster}

The first force based surrogate predicts the objective at correlation $0.9996$ while its
gradient correlation reaches only $0.459$, so it was discarded before any controller was built.
The field based latent surrogate passes the held-out audit on 72 gradient labels at $0.938$,
$0.968$, and $0.908$. A second held-out audit on 48 comparisons from 16 additional states gives $0.936$, $0.938$, and
$0.638$.

Two deployment protocols were run. The core protocol uses three perturbed initial conditions
and 240 scored steps each, where the extracted law, the anchor-held policy, and the
anchor-removed policy give $4.60$, $5.33$, and $-2.98\%$. The frozen robustness protocol uses eight initial
conditions and 600 scored steps each, where one training seed gives $6.71$, $9.76$, and $9.66\%$. Therefore, the sign of the reduction of the released controller depends on the horizon and on the set of initial conditions, and a benefit of release that holds under both protocols is not supported. The anchor-removed controller uses more actuation under
both protocols, with mean action norm $1.29$ and $1.35$ against $0.764$ and $0.832$ for the
anchor-held policy and $0.521$ for the extracted law. The state-independent part of this flow's measured gradient field holds
$83.8\%$ of its energy, with a resampling stability interval from $0.808$ to $0.903$.

\subsection{Forced two-dimensional turbulence}

The cone-residual controller is frozen before training seeds 101 through 108 are run. Its action
is constrained to a five degree cone around the audited direction. Every training seed passes the controller check, with median cosine to the high-fidelity descent direction between $0.9893$ and $0.9901$.
The eight training seed mean reductions range from $81.77$ to $82.18\%$. Their mean is
$82.01\pm0.19\%$, compared with $30.63\pm21.43\%$ for PPO over four training seeds. The advantage of $51.38$ percentage points has a Welch $95\%$ confidence interval from $17.28$ to $85.48$.

\subsection{Parametric chaotic equation}

At $\mu=0.225$, the surrogate-based controller costs $361.58\pm8.31$, a sample standard deviation, on eight held-out
fields. Its gradient model, which predicts the action gradient over 50 steps, reaches pooled correlation $0.979$ on the held-out audit states and $0.956$ on states reached by an earlier controller built on the same gradient model. The controller keeps the
audited gradient as the direction of its action while using the available actuator
authority. The comparison detects no difference from matched plain TD3, and the controller is lower in cost than matched
polynomial TD3 on every confirmation field.

At $\mu=0.121$, a predefined diagnostic tests the difference between the two forcing amplitudes under the benchmark observation
noise. The noise standard deviation is $0.25$. Under the transferred controller, the clean
controlled sensor root mean square divided by this noise level is $0.959$ at $\mu=0.121$ and
$2.05$ at $\mu=0.225$. Therefore, the lower amplitude puts the measured signal at the noise
floor. Across matched sources $5401$ through $5404$, adding noise at $\mu=0.121$ raises the raw
transferred controller cost by $71.06$, or $109.1\%$. All five predefined mechanism rules
pass.

A controlled linear one-step observer is fitted on $3{,}992$ clean transitions. Ridge coefficient
$0.0001$ gives held-out source correlation $0.9991$ and root mean square error $0.0096$. Frozen
selection on sources $5411$ through $5414$ chooses an observer that keeps $10\%$ of the new
measurement and $90\%$ of the controlled prediction.

The observer is then frozen and evaluated on held-out sources $5501$ through $5508$. Its cost
is $74.92\pm4.01$, compared with $71.93$ for matched plain TD3 and $83.75$ for matched polynomial
TD3. The paired observer minus plain TD3 difference
is $+2.99$, with a $95\%$ interval from $-1.09$ to $+7.08$. The upper limit lies below the fixed
$10\%$ noninferiority margin of $7.19$. The observer minus polynomial TD3 difference is
$-8.83$, with an interval from $-11.74$ to $-5.91$, so the controller with observer is better on this
comparison. All predefined confirmation rules pass.

\end{document}